\documentclass{article}
\usepackage{ijcai21}

\usepackage{amsmath,amsfonts,bm}

\def\eqref#1{equation~\ref{#1}}

\def\1{\bm{1}}

\def\vp{{\bm{p}}}

\def\vr{{\bm{r}}}

\def\vx{{\bm{x}}}
\def\vy{{\bm{y}}}
\def\vz{{\bm{z}}}

\DeclareMathAlphabet{\mathsfit}{\encodingdefault}{\sfdefault}{m}{sl}
\SetMathAlphabet{\mathsfit}{bold}{\encodingdefault}{\sfdefault}{bx}{n}

\usepackage{times}
\usepackage{soul}
\usepackage{url}
\usepackage[hidelinks]{hyperref}
\usepackage[utf8]{inputenc}
\usepackage[small]{caption}
\usepackage{graphicx}
\usepackage{amssymb,mathtools}
\usepackage{amsmath}
\usepackage{amsthm}
\usepackage{booktabs}
\usepackage{algorithmic}
\usepackage{xcolor}
\usepackage{floatrow}
\usepackage[english]{babel}
\usepackage{subcaption}
\usepackage{mathtools}
\usepackage{multirow}
\usepackage{wrapfig}
\usepackage{amssymb}
\usepackage{bbm}
\usepackage{makecell}

\newtheorem{proposition}{Proposition}

\title{Comparing Kullback-Leibler Divergence and Mean Squared Error Loss\\
in Knowledge Distillation}

\author{
Taehyeon Kim\thanks{The authors contributed equally to this paper.}$^{1}$\and
Jaehoon Oh\footnotemark[1]$^{2}$\and
Nak Yil Kim$^1$\and
Sangwook Cho$^1$\And
Se-Young Yun$^1$
\affiliations
$^1$Graduate School of Artificial Intelligence, KAIST\\
$^2$Graduate School of Knowledge Service Engineering, KAIST
\emails
\{potter32, jhoon.oh, nakyilkim, sangwookcho, yunseyoung\}@kaist.ac.kr
}

\begin{document}

\maketitle

\begin{abstract}
Knowledge distillation (KD), transferring knowledge from a cumbersome teacher model to a lightweight student model, has been investigated to design efficient neural architectures. Generally, the objective function of KD is the Kullback-Leibler (KL) divergence loss between the softened probability distributions of the teacher model and the student model with the temperature scaling hyperparameter $\tau$. Despite its widespread use, few studies have discussed the influence of such softening on generalization. Here, we theoretically show that the KL divergence loss focuses on the \emph{logit matching} when $\tau$ increases and the \emph{label matching} when $\tau$ goes to 0 and empirically show that the logit matching is positively correlated to performance improvement in general. From this observation, we consider an intuitive KD loss function, the mean squared error (MSE) between the logit vectors, so that the student model can directly learn the logit of the teacher model. The MSE loss outperforms the KL divergence loss, explained by the difference in the penultimate layer representations between the two losses. Furthermore, we show that sequential distillation can improve performance and that KD, particularly when using the KL divergence loss with small $\tau$, mitigates the label noise. The code to reproduce the experiments is publicly available online at {\color{blue} \url{https://github.com/jhoon-oh/kd_data/}}.

\end{abstract}

\section{Introduction}\label{sec:intro}

Knowledge distillation\,(KD) is one of the most potent model compression techniques in which knowledge is transferred from a cumbersome model\,(teacher) to a single small model\,(student)\,\cite{hinton2015distilling}. In general, the objective of training a smaller student network in the KD framework is formed as a linear summation of two losses: cross-entropy\,(CE) loss with ``hard'' targets, which are one-hot ground-truth vectors of the samples, and Kullback-Leibler\,(KL) divergence loss with the teacher's predictions. Specifically, KL divergence loss has achieved considerable success by controlling the softness of ``soft'' targets via the temperature-scaling hyperparameter $\tau$. Utilizing a larger value for this hyperparameter $\tau$ makes the softmax vectors smooth over classes. Such a re-scaled output probability vector by $\tau$ is called the softened probability distribution, or the softened softmax \cite{DBLP:journals/corr/MaddisonMT16,jang2016categorical}. Recent KD has evolved to give more importance to the KL divergence loss to improve performance when balancing the objective between CE loss and KL divergence loss\,\cite{hinton2015distilling,tian2019contrastive}. Hence, we focus on training a student network based solely on the KL divergence loss.




\begin{figure}[t!]
    \centering
    \includegraphics[width=\linewidth]{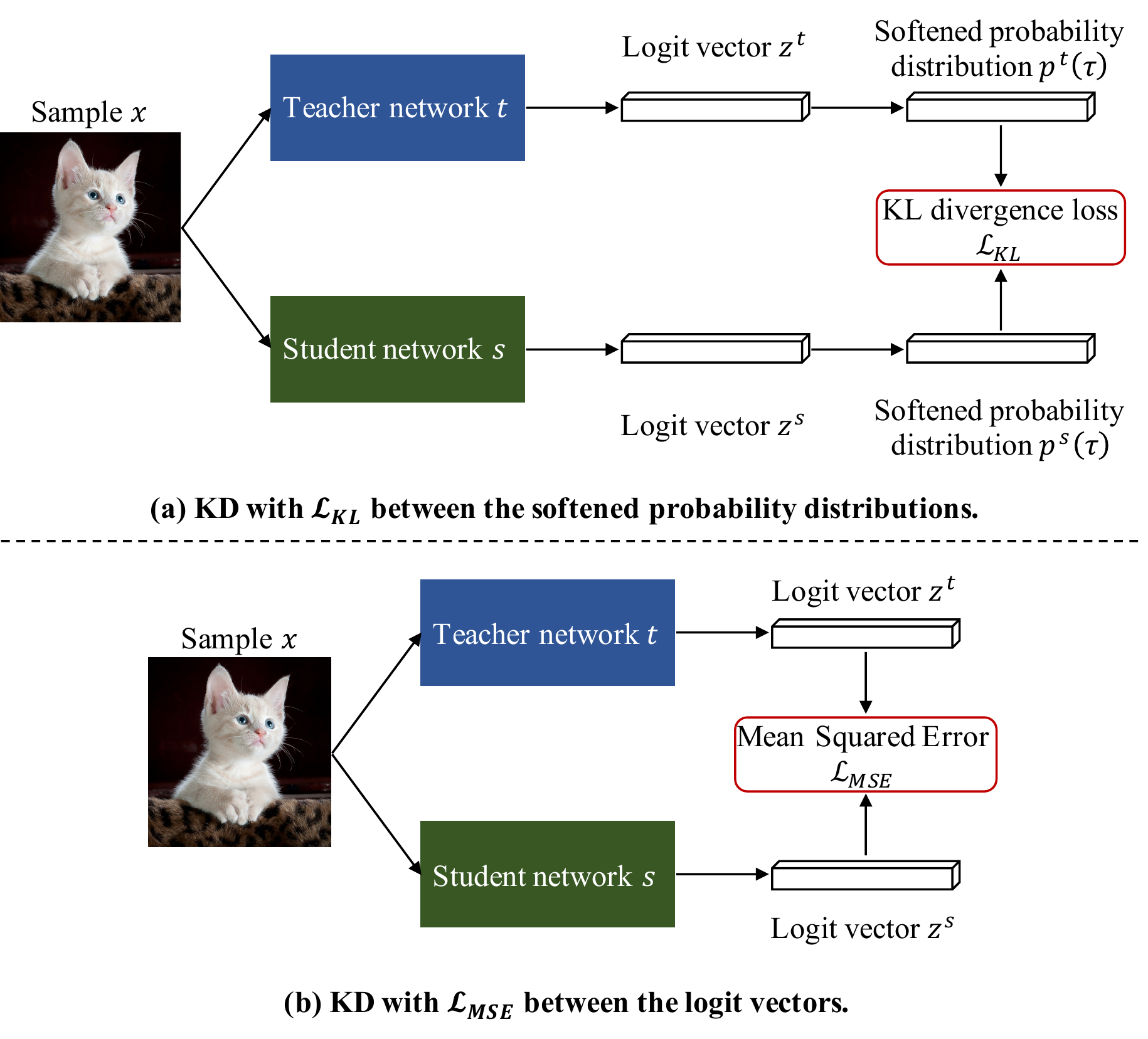}
    \caption{Overview of knowledge distillation (KD) without the CE loss $\mathcal{L}_{CE}$ with a ground-truth vector: KD framework with (a) KL divergence loss $\mathcal{L}_{KL}$ and (b) mean squared error loss $\mathcal{L}_{MSE}$.}
    \label{fig:overview}
\end{figure}

Recently, there has been an increasing demand for investigating the reasons for the superiority of KD.
\cite{yuan2020revisit,tang2020understanding} empirically showed that the facilitation of KD is attributed to not only the privileged information on similarities among classes but also the label smoothing regularization. In some cases, the theoretical reasoning for using ``soft'' targets is clear. For example, in deep linear neural networks, KD not only accelerates the training convergence but also helps in reliable training\,\cite{phuong2019towards}. In the case of self-distillation\,(SD), where the teacher model and the student model are the same, such an approach progressively restricts the number of basis functions to represent the solution\,\cite{mobahi2020self}. However, there is still a lack of understanding of how the degree of softness affects the performance.


In this paper, we first investigate the characteristics of a student trained with KL divergence loss with various $\tau$, both theoretically and empirically. We find that the student's logit\,(i.e., an input of the softened softmax function) more closely resembles the teacher's logit as $\tau$ increases, but not completely. Therefore, we design a direct logit learning scheme by replacing the KL divergence loss between the softened probability distributions of a teacher network and a student network (\autoref{fig:overview}(a)) with the mean squared error\,(MSE) loss between the student's logit and the teacher's logit (\autoref{fig:overview}(b)). Our contributions are summarized as follows:

\begin{itemize}
    \item We investigate the role of the softening hyperparameter $\tau$ theoretically and empirically. A large $\tau$, that is, strong softening, leads to \emph{logit matching}, whereas a small $\tau$ results in training \emph{label matching}. In general, \emph{logit matching} has a better generalization capacity than \emph{label matching}.
    \item We propose a direct logit matching scheme with the MSE loss and show that the KL divergence loss with any value of $\tau$ cannot achieve complete logit matching as much as the MSE loss. Direct training results in the best performance in our experiments.
    \item We theoretically show that the KL divergence loss makes the model's penultimate layer representations elongated than those of the teacher, while the MSE loss does not. We visualize the representations using the method proposed by \cite{muller2019does}.
    \item We show that sequential distillation, the MSE loss after the KL divergence loss, can be a better strategy than direct distillation when the capacity gap between the teacher and the student is large, which contrasts \cite{cho2019efficacy}.
    \item We observe that the KL divergence loss, with low $\tau$ in particular, is more efficient than the MSE loss when the data have incorrect labels\,(\textit{noisy label}). In this situation, extreme logit matching provokes bad training, whereas the KL divergence loss mitigates this problem.
\end{itemize}

\section{Related Work}

\subsection{Knowledge Distillation}
KD has been extended to a wide range of methods. One attempted to distill not only the softened probabilities of the teacher network but also the hidden feature vector so that the student could be trained with rich information from the teacher \cite{romero2014fitnets,zagoruyko2016paying,srinivas2018knowledge,kim2018paraphrasing,heo2019knowledge,heo2019comprehensive}. The KD approach can be leveraged to reduce the generalization errors in teacher models (i.e., self-distillation; SD)\,\cite{Zhang_2019_ICCV,park2019relational} as well as model compressions. In the generative models, a generator can be compressed by distilling the latent features from a cumbersome generator \cite{DBLP:journals/corr/abs-1902-00159}.

To explain the efficacy of KD, \cite{furlanello2018born} asserted that the maximum value of a teacher's softmax probability was similar to weighted importance by showing that permuting all of the non-argmax elements could also improve performance. \cite{yuan2020revisit} argued that ``soft'' targets served as a label smoothing regularizer rather than as a transfer of class similarity by showing that a poorly trained or smaller-size teacher model can boost performance. Recently, \cite{tang2020understanding} modified the conjecture in \cite{furlanello2018born} and showed that the sample was positively re-weighted by the prediction of the teacher's logit vector.

\subsection{Label Smoothing}
Smoothing the label $\vy$ is a common method for improving the performance of deep neural networks by preventing the overconfident predictions\,\cite{szegedy2016rethinking}. Label smoothing is a technique that facilitates the generalization by replacing a ground-truth one-hot vector $\vy$ with a weighted mixture of hard targets $\vy^{LS}$:
\begin{equation}
    \vy^{LS}_k =  
    \begin{cases}
    (1-\beta)       \quad\text{if $\vy_k=1$} \\
    \frac{\beta}{K-1} \quad\quad\text{otherwise}\\
    \end{cases}
\end{equation}

\noindent where $k$ indicates the index, and $\beta$ is a constant. This implicitly ensures that the model is well-calibrated\,\cite{muller2019does}. Despite its improvements, \cite{muller2019does} observed that the teacher model trained with LS improved its performance, whereas it could hurt the student's performance. \cite{yuan2020revisit} demonstrated that KD might be a category of LS by using the adaptive noise, i.e., KD is a label regularization method.


\section{Preliminaries: KD}\label{sec:prelim_KD}


\begin{table}
\centering
\small
\begin{tabular}{c|l}
    \toprule
    Notation & Description \\
    \midrule \midrule
    $\vx$                           & Sample \\ \midrule
    $\vy$                           & Ground-truth one-hot vector \\ \midrule
    $K$                             & Number of classes in the dataset \\ \midrule
    $f$                             & Neural network    \\ \midrule
    $\vz^f$                    & Logit vector of a sample $\vx$ through a network $f$ \\ \midrule
    \multirow{2}{*}{$\vz^f_k$} & Logit value corresponding the $k$-th class label, i.e.,\\
                                    & the $k$-th value of $z^f(\vx)$ \\ \midrule
    $\alpha$                        & Hyperparameter of the linear combination  \\ \midrule
    $\tau$                          & Temperature-scaling hyperparameter \\ \midrule
    \multirow{2}{*}{$\vp^{f}(\tau)$}     & Softened probability distribution with $\tau$ of a sample $\vx$ \\
                                              & for a network $f$ \\ \midrule
    \multirow{3}{*}{$\vp^{f}_{k}(\tau)$} & The $k$-th value of a softened probability distribution, \\ 
                                              & i.e.,  $ \frac{exp({\vz^f_k/\tau})} {\sum_{j=1}^{K} exp({\vz^f_j/\tau})} $ \\ \midrule
    $\mathcal{L}_{CE}$  & Cross-entropy loss    \\ \midrule
    $\mathcal{L}_{KL}$  & Kullback-Leibler divergence loss    \\ \midrule
    $\mathcal{L}_{MSE}$ & Mean squared error loss \\ \bottomrule
\end{tabular}
\caption{Mathematical terms and notations in our work.}
\label{tab:notation}
\end{table}

We denote the softened probability vector with a temperature-scaling hyperparameter $\tau$ for a network $f$ as $\vp^{f}(\tau)$, given a sample $\vx$. The $k$-th value of the softened probability vector $\vp^{f}(\tau)$ is denoted by $\vp^{f}_{k}(\tau)= \frac {\exp({\vz^f_k/\tau})}{\sum_{j=1}^{K} \exp({\vz^f_j/\tau})}$, where $\vz^f_k$ is the $k$-th value of the logit vector $\vz^f$, $K$ is the number of classes, and $\exp(\cdot)$ is the natural exponential function. Then, given a sample $\vx$, the typical loss $\mathcal{L}$ for a student network is a linear combination of the cross-entropy loss $\mathcal{L}_{CE}$ and the Kullback-Leibler divergence loss $\mathcal{L}_{KL}$:

\begin{equation}\label{eq:KD_loss}
\resizebox{0.91\hsize}{!}{%
$\begin{gathered}
    \mathcal{L} = (1-\alpha) \mathcal{L}_{CE}(\vp^{s}(1), \vy) + \alpha \mathcal{L}_{KL}(\vp^{s}(\tau), \vp^{t}(\tau)), \\
    \mathcal{L}_{CE}(\vp^{s}(1), \vy) = \sum_{j} - \vy_j \log \vp^{s}_j(1) \\ 
    \mathcal{L}_{KL}(\vp^{s}(\tau), \vp^{t}(\tau)) = \tau^2 \sum_{j} \vp^{t}_j(\tau) \log \frac{\vp^{t}_j(\tau)}{\vp^{s}_j(\tau)}
\end{gathered}$
}
\end{equation}

\noindent where $s$ indicates the student network, $t$ indicates the teacher network, $\vy$ is a one-hot label vector of a sample $\vx$, and $\alpha$ is a hyperparameter of the linear combination. For simplicity of notation, $\mathcal{L}_{CE}(\vp^{s}(1), \vy)$ and $\mathcal{L}_{KL}(\vp^{s}(\tau), \vp^{t}(\tau))$ are denoted by $\mathcal{L}_{CE}$ and $\mathcal{L}_{KL}$, respectively. The standard choices are $\alpha = 0.1$ and $\tau \in \{3,4,5\}$ \cite{hinton2015distilling,zagoruyko2016paying}.

In \cite{hinton2015distilling}, given a single sample $\vx$, the gradient of $\mathcal{L}_{KL}$ with respect to $\vz^s_k$ is as follows:

\begin{equation}\label{eq:KL_grad}
    \frac{\partial\mathcal{L}_{KL}}{\partial{\vz^s_k}} = \tau (\vp^{s}_{k}(\tau) - \vp^{t}_{k}( \tau))
\end{equation}

\noindent When $\tau$ goes to $\infty$, this gradient is simplified with the approximation, i.e., $exp(\vz^f_k/\tau) \approx 1+\vz^f_k/\tau$:
\begin{equation}\label{eq:KL_grad_approx}
    \frac{\partial\mathcal{L}_{KL}}{\partial{\vz^s_k}} \approx \tau \left( \frac{1 + \vz^s_k/\tau}{K + \sum_{j} \vz^s_j/\tau} - \frac{1 + \vz^t_k/\tau}{K + \sum_{j} \vz^t_j/\tau} \right)
\end{equation}


\noindent Here, the authors assumed the zero-mean teacher and student logit, i.e., $\sum_{j} \vz^t_j = 0$ and $\sum_{j} \vz^s_j = 0$, and hence $\frac{\partial\mathcal{L}_{KL}}{\partial{\vz^s_k}} \approx \frac{1}{K} (\vz^s_k - \vz^t_k)$. This indicates that minimizing $\mathcal{L}_{KL}$ is equivalent to minimizing the mean squared error $\mathcal{L}_{MSE}$, that is, $ || \vz^s-\vz^t ||^{2}_{2}$, under a sufficiently large temperature $\tau$ and the zero-mean logit assumption for both the teacher and the student.

\begin{figure}[t]
    \begin{minipage}{.48\textwidth}
    \centering
    \includegraphics[width=\linewidth]{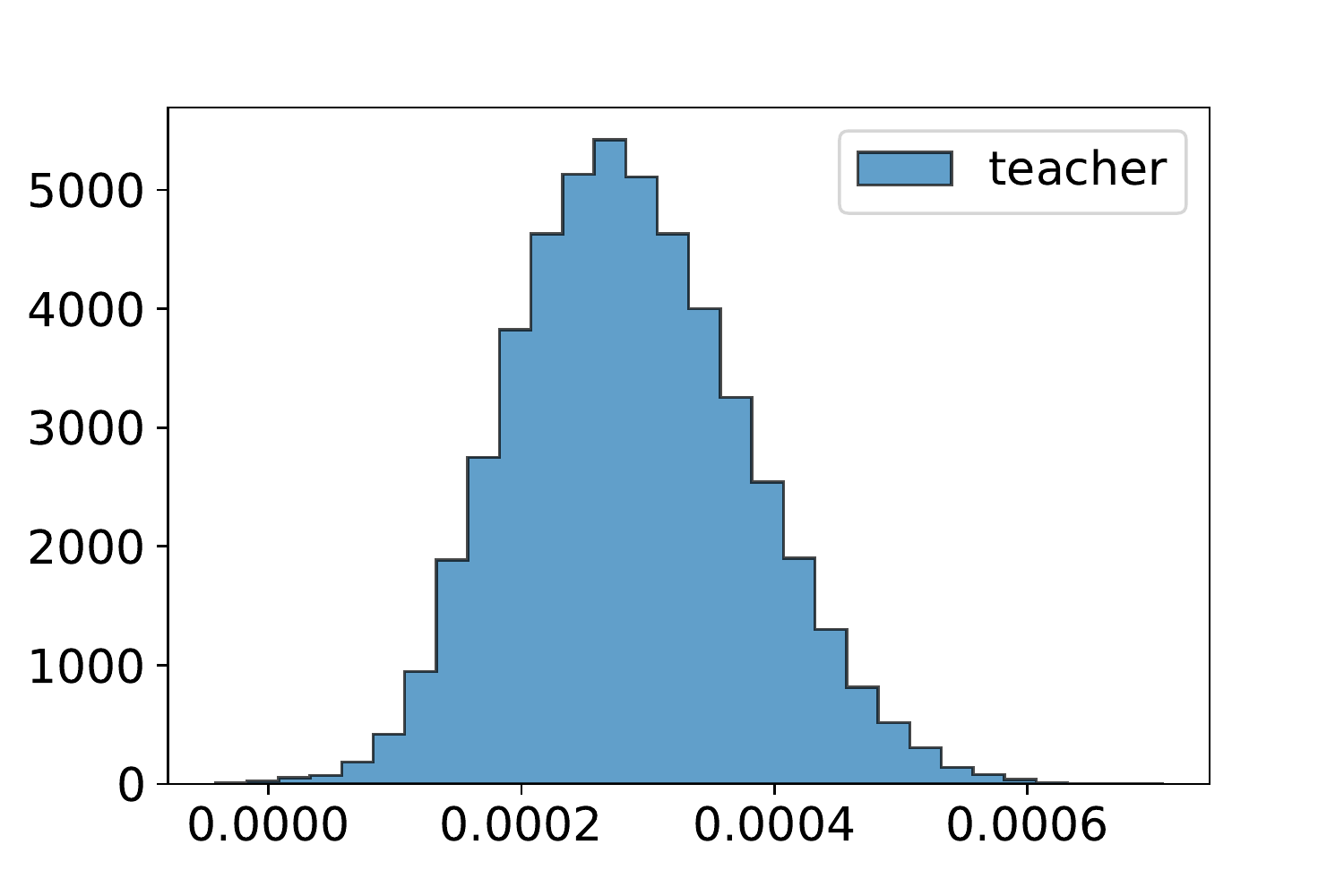}
    \subcaption{Teacher.}
    \label{fig:tea_summation}
    \end{minipage}\hfill
    \begin{minipage}{.48\textwidth}
    \centering
    \includegraphics[width=\linewidth]{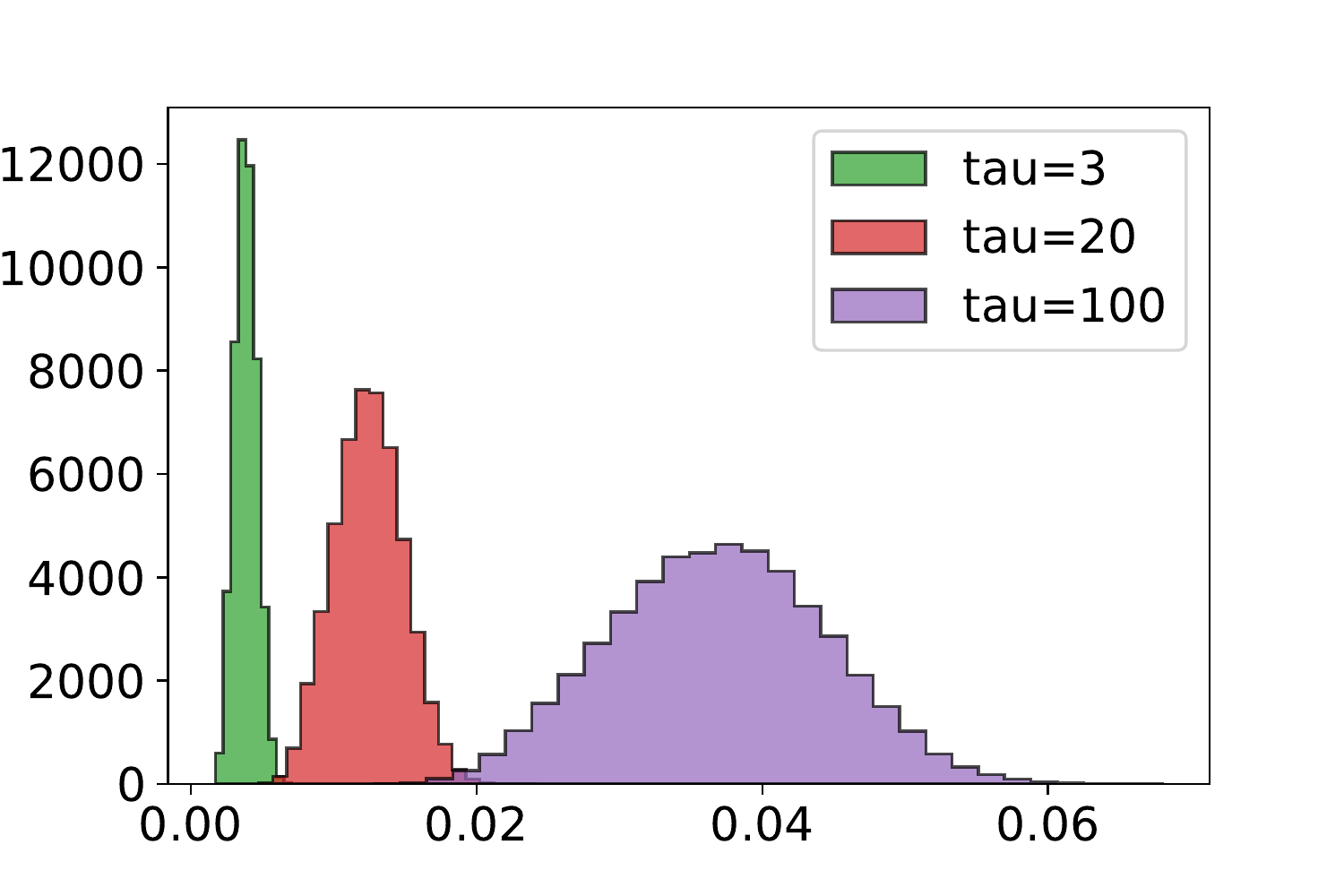}
    \subcaption{Student with $\mathcal{L}_{KL}.$}
    \end{minipage} \hfill
\caption{Histograms for the magnitudes of logit summation on the CIFAR-100 training dataset. We use a (teacher, student) pair of (WRN-28-4, WRN-16-2).} \label{fig:prelim_logit_sum}
\end{figure}

However, we observe that this assumption does not seem appropriate and hinders complete understanding by ignoring the hidden term in $\mathcal{L}_{KL}$ when $\tau$ increases. \autoref{fig:prelim_logit_sum} describes the histograms for the magnitude of logit summations on the training dataset. The logit summation histogram from the teacher network trained with $\mathcal{L}_{CE}$ is almost zero (\autoref{fig:prelim_logit_sum}(a)), whereas that from the student network trained with $\mathcal{L}_{KL}$ using the teacher's knowledge goes far from zero as $\tau$ increases (\autoref{fig:prelim_logit_sum}(b)). This is discussed in detail in Section \ref{sec:KL_to_MSE}.

\subsection{Experimental Setup}
In this paper, we used an experimental setup similar to that in \cite{heo2019comprehensive,cho2019efficacy}: image classification on CIFAR-100 with a family of Wide-ResNet\,(WRN) \cite{zagoruyko2016wide} and ImageNet with a family of of ResNet\,(RN)\,\cite{he2016deep}. We used a standard PyTorch SGD optimizer with a momentum of 0.9, weight decay, and apply standard data augmentation. Other than those mentioned, the training settings from the original papers\,\cite{heo2019comprehensive,cho2019efficacy} were used.


\section{Relationship between $\mathcal{L}_{KL}$ and $\mathcal{L}_{MSE}$}\label{sec:tau}

In this section, we conduct extensive experiments and systematically break down the effects of $\tau$ in $\mathcal{L}_{KL}$ based on theoretical and empirical results. Then, we highlight the relationship between $\mathcal{L}_{KL}$ and $\mathcal{L}_{MSE}$. Then, we compare the models trained with $\mathcal{L}_{KL}$ and $\mathcal{L}_{MSE}$ in terms of performance and penultimate layer representations. Finally, we investigate the effects of a noisy teacher on the performance according to the objective.

\subsection{Hyperparameter $\tau$ in $\mathcal{L}_{KL}$}

\begin{figure}[t]
\centering
    \begin{minipage}{1.0\linewidth}
        \centering
        \begin{minipage}[t]{.49\linewidth}
        \centering
        \includegraphics[width=\linewidth]{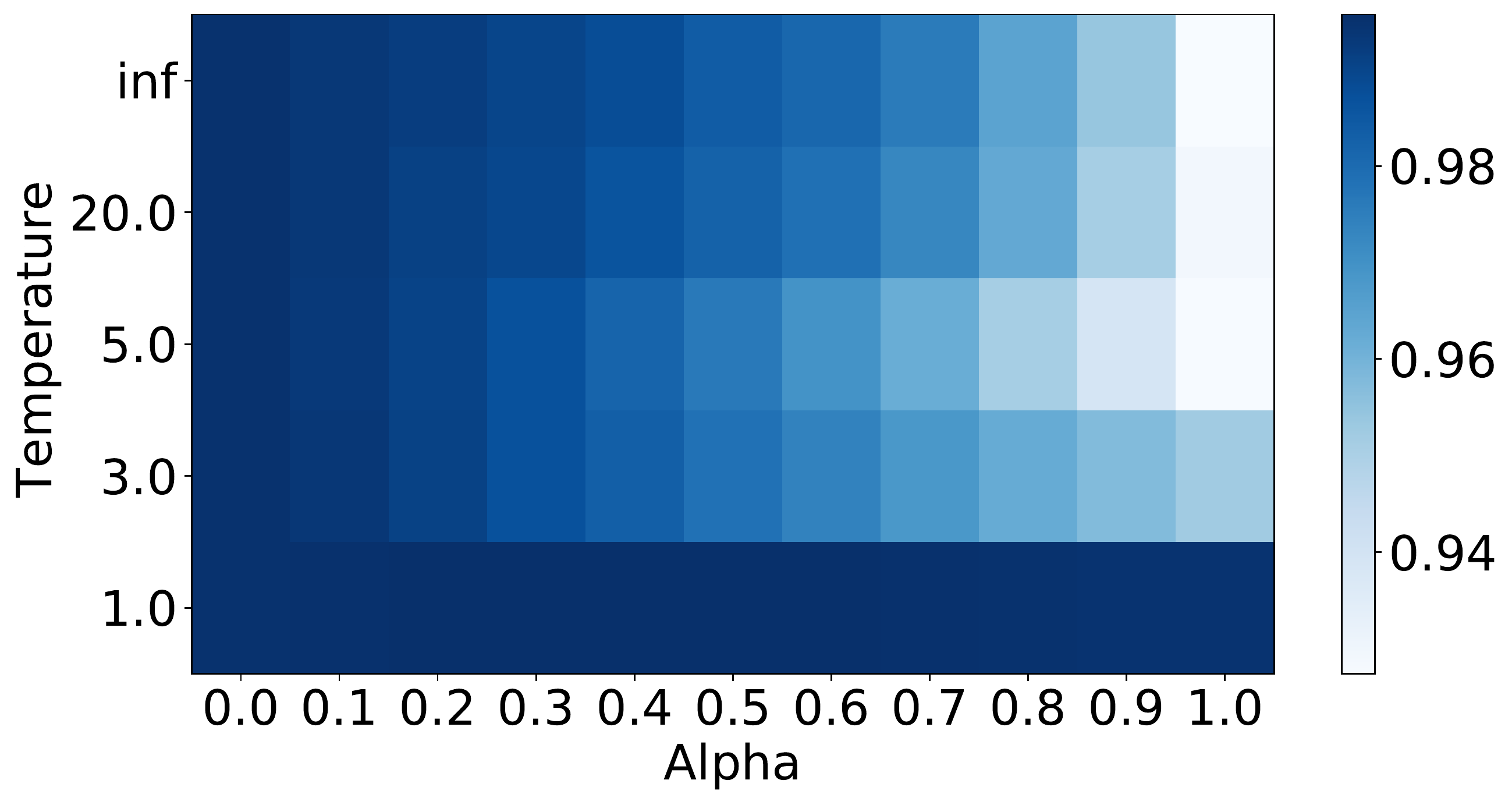}
        \subcaption{Training accuracy.}\label{fig:train_acc}
        \end{minipage}
        \begin{minipage}[t]{.49\linewidth}
        \centering
        \includegraphics[width=\linewidth]{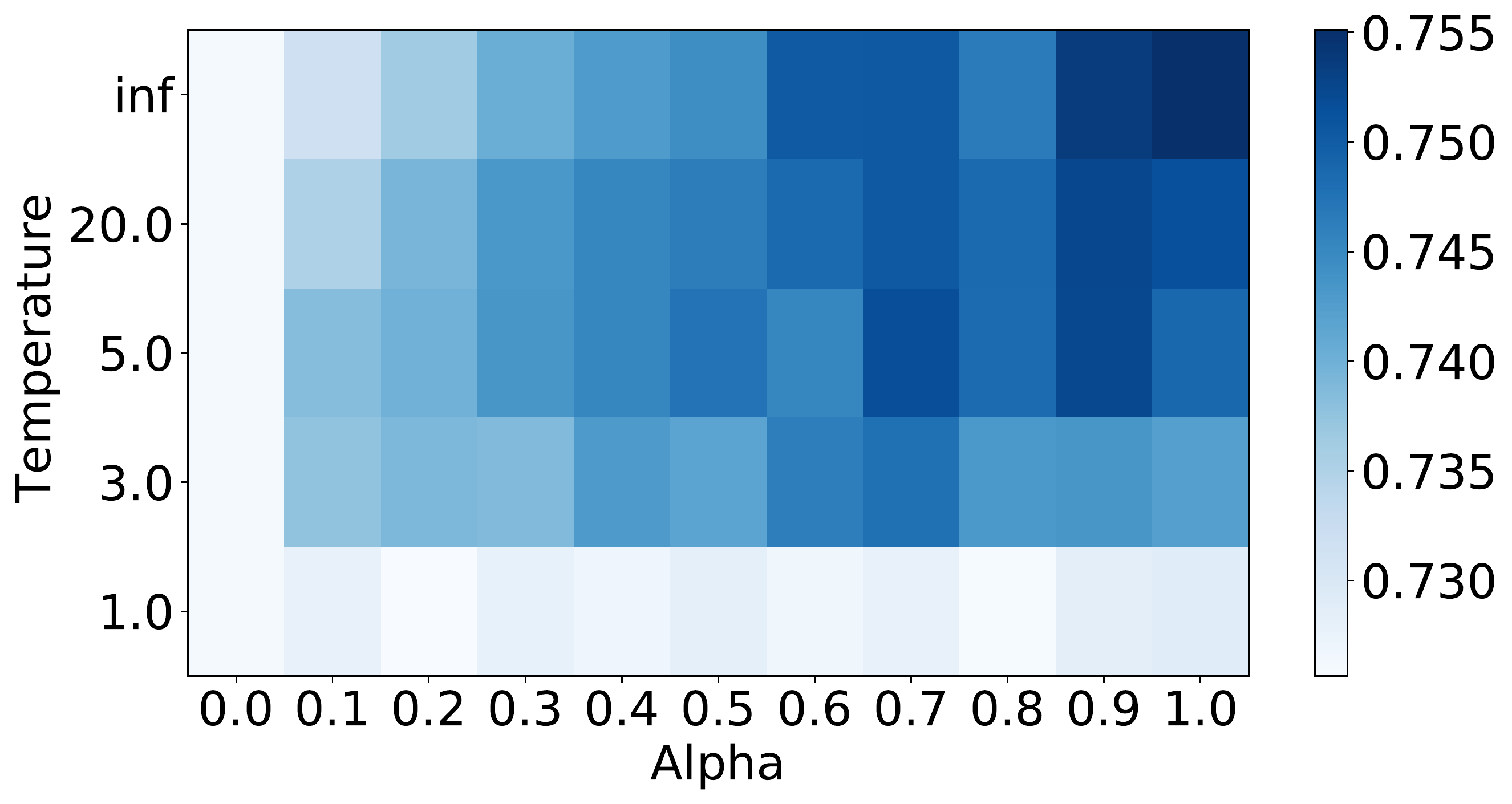}
        \subcaption{Test accuracy.}\label{fig:test_acc}
        \end{minipage}
    \end{minipage}
    \caption{\textbf{Grid maps of accuracies according to the change of $\alpha$ and $\tau$} on CIFAR-100 when (teacher, student) = (WRN-28-4, WRN-16-2). It presents the grid maps of (a) \textbf{training top-1 accuracies} and (b) \textbf{test top-1 accuracies}. $\mathcal{L}_{KL}$ with $\tau=\infty$ is implemented using a handcrafted gradient (Eq.~(\ref{eq:to_inf}))}\label{fig:accuracy}
\end{figure}

We investigate the training and test accuracies according to the change in $\alpha$ in $\mathcal{L}$ and $\tau$ in $\mathcal{L}_{KL}$ (\autoref{fig:accuracy}). First, we empirically observe that the generalization error of a student model decreases as $\alpha$ in $\mathcal{L}$ increases. This means that ``soft'' targets are more efficient than ``hard'' targets in training a student if ``soft'' targets are extracted from a well-trained teacher. This result is consistent with prior studies that addressed the efficacy of ``soft'' targets\,\cite{furlanello2018born,tang2020understanding}. Therefore, we focus on the situation where ``soft'' targets are used to train a student model solely, that is, $\alpha = 1.0$, in the remainder of this paper.

When $\alpha=1.0$, the generalization error of the student model decreases as $\tau$ in $\mathcal{L}_{KL}$ increases. These consistent tendencies according to the two hyperparameters, $\alpha$ and $\tau$, are the same across various teacher-student pairs. To explain this phenomenon, we extend the gradient analysis in Section \ref{sec:prelim_KD} without the assumption that the mean of the logit vector is zero.



\begin{proposition}\label{thm1}
Let $K$ be the number of classes in the dataset, and $\1{[\cdot]}$ be the indicator function, which is 1 when the statement inside the brackets is true and 0 otherwise. Then,
\begin{equation}\label{eq:to_inf}
\begin{split}
    \lim_{\tau \to \infty}\frac{\partial\mathcal{L}_{KL}}{\partial{\vz^s_k}} & = \frac{1}{K^2} \sum_{j=1}^K \left( (\vz^s_k - \vz^s_j) - (\vz^t_k - \vz^t_j) \right) \\
    & = \frac{1}{K} \left( \vz^s_k - \vz^t_k \right) - \frac{1}{K^2} \sum_{j=1}^K \left( \vz^s_j - \vz^t_j \right)
\end{split}
\end{equation}
\begin{equation}\label{eq:to_zero}
    \lim_{\tau \to 0}\frac{1}{\tau}\frac{\partial\mathcal{L}_{KL}}{\partial{\vz^s_k}} = \1_{[ \arg\max_j \vz^s_j = k]} - \1_{[\arg\max_j \vz^t_j = k]} 
\end{equation}
\end{proposition}

Proposition\,\autoref{thm1} explains the consistent trends as follows. In the course of regularizing $\mathcal{L}_{KL}$ with sufficiently large $\tau$, the student model attempts to imitate the logit distribution of the teacher model. Specifically, a larger $\tau$ is linked to a larger $\mathcal{L}_{KL}$, making the logit vector of the student similar to that of the teacher\,(i.e.,\,\emph{logit matching}). Hence, ``soft'' targets are being fully used as $\tau$ increases. This is implemented using a handcrafted gradient\,(top row of \autoref{fig:accuracy}).
On the other hand, when $\tau$ is close to 0, the gradient of $\mathcal{L}_{KL}$ does not consider the logit distributions and only identifies whether the student and the teacher share the same output (i.e., \emph{label matching}), which transfers limited information. In addition, there is a scaling issue when $\tau$ approaches 0. As $\tau$ decreases, $\mathcal{L}_{KL}$ increasingly loses its quality and eventually becomes less involved in learning. The scaling problem can be easily fixed by multiplying $1/\tau$ by $\mathcal{L}_{KL}$ when $\tau$ is close to zero.

From this proposition, it is recommended to modify the original $\mathcal{L}_{KL}$ in Eq.~(\ref{eq:KD_loss}), considering $\tau \in (0, \infty)$ as follows:

\begin{equation}\label{eq:KL_modification}
\begin{gathered}
    \max({\tau, \tau^2}) \sum_{j} \vp^{t}_j(\tau) \log \frac{\vp^{t}_j(\tau)}{\vp^{s}_j(\tau)}
\end{gathered}
\end{equation}

The key difference between our analysis and the preliminary analysis on a sufficiently large $\tau$, i.e., in Eq.~(\ref{eq:to_inf}), is that the latter term is generated by removing the existing assumption on the logit mean, which is discussed in Section \ref{sec:KL_to_MSE} at the loss-function level.

\subsection{Extensions from $\mathcal{L}_{KL}$ to $\mathcal{L}_{MSE}$}\label{sec:KL_to_MSE}
In this subsection, we focus on Eq.~(\ref{eq:to_inf}) to investigate the reason as to why the efficacy of KD is observed when $\tau$ is greater than 1 in the KD environment, as shown in \autoref{fig:accuracy}. Eq.~(\ref{eq:to_inf}) can be understood as a biased regression of the vector expression as follows:
\begin{gather*}\label{eq:to_inf2}
    \lim_{\tau \to \infty} \nabla_{\vz^s} \mathcal{L}_{KL} = \frac{1}{K} \left( \vz^s - \vz^t \right ) - \frac{1}{K^2} \sum_{j=1}^K \left( \vz^s_j - \vz^t_j \right ) \cdot \mathbbm{1}
\end{gather*}
where $\mathbbm{1}$ is a vector whose elements are equal to one. Furthermore, we can derive the relationship between $\lim_{\tau \to \infty} \mathcal{L}_{KL}$ and $\mathcal{L}_{MSE}$ as follows:

\begin{equation}
\resizebox{0.91\hsize}{!}{%
$\begin{gathered}
     \lim_{\tau \to \infty} \mathcal{L}_{KL} = \frac{1}{2K} || \vz^s - \vz^t ||_2^2  + \delta_{\infty} = \frac{1}{2K} \mathcal{L}_{MSE} + \delta_{\infty} \\
    \delta_{\infty} = - \frac{1}{2K^2} (\sum_{j=1}^K \vz_j^s - \sum_{j=1}^K \vz_j^t)^2 + Constant\\
\end{gathered}$
}\label{eq:relation}
\end{equation}


In  \autoref{fig:prelim_logit_sum}(a), the sum of the logit values of the teacher model is almost zero. With the teacher's logit value,  $\delta_{\infty}$ is approximated as $ - \frac{1}{2K^2} ({\sum_{j=1}^K \vz_j^s})^2$. Therefore, $\delta_{\infty}$ can make the logit mean of the student trained with $\mathcal{L}_{KL}$ depart from zero. From this analysis, it is unreasonable to assume that the student's logit mean is zero. We empirically find that the student's logit mean breaks the existing assumption as $\tau$ increases\,(\autoref{fig:prelim_logit_sum}(b)). In summary, \emph{$\delta_\infty$ hinders complete logit matching by shifting the mean of the elements in the logit}. In other words, as derived from Eq.~(\ref{eq:relation}), optimizing $\mathcal{L}_{KL}$ with sufficiently large $\tau$ is equivalent to optimizing $\mathcal{L}_{MSE}$ with the additional regularization term $\delta_\infty$, and it seems to rather hinder logit matching. 

Therefore, we propose the direct logit learning objective for enhanced logit matching as follows:

\begin{equation}\label{eq:new_KD_loss}
\begin{gathered}
    \mathcal{L}^\prime = (1-\alpha) \mathcal{L}_{CE}(\vp^{s}(1), \vy) + \alpha \mathcal{L}_{MSE}(\vz^{s}, \vz^{t}), \\
    \mathcal{L}_{MSE}(\vz^{s}, \vz^{t}) = || \vz^s - \vz^t ||_2^2
\end{gathered}
\end{equation}

\noindent Although this direct logit learning was used in \cite{ba2013deep,urban2016deep}, they did not investigate the wide range of temperature scaling and the effects of MSE in the latent space. In this respect, our work differs.

\subsection{Comparison of $\mathcal{L}_{KL}$ and $\mathcal{L}_{MSE}$}

\begin{table}[t]
\centering
\scriptsize
\begin{tabular}{c|ccccccc}
    \toprule
    \multirow{2}{*}{Student} & \multirow{2}{*}{$\mathcal{L}_{CE}$} & \multicolumn{5}{c}{$\mathcal{L}_{KL}$} & \multirow{2}{*}{$\mathcal{L}_{MSE}$} \\
    \cmidrule{3-7}
     &  & $\tau$=1 & $\tau$=3 & $\tau$=5 & $\tau$=20 & $\tau$=$\infty$ &  \\
    \midrule\midrule
    WRN-16-2 & 72.68 & 72.90 & 74.24 & 74.88 & 75.15 & 75.51 & \textbf{75.54} \\
    WRN-16-4 & 77.28 & 76.93 & 78.76 & 78.65 & 78.84 & 78.61 & \textbf{79.03} \\
    WRN-28-2 & 75.12 & 74.88 & 76.47 & 76.60 & \textbf{77.28} & 76.86 & \textbf{77.28} \\
    WRN-28-4 & 78.88 & 78.01 & 78.84 & 79.36 & 79.72 & 79.61 & \textbf{79.79} \\
    WRN-40-6 & 79.11 & 79.69 & 79.94 & 79.87 & 79.82 & 79.80 & \textbf{80.25} \\
    \bottomrule
\end{tabular}
\caption{Top-1 test accuracies on CIFAR-100. WRN-28-4 is used as a teacher for $\mathcal{L}_{KL}$ and $\mathcal{L}_{MSE}$.}
\label{tab:mse_loss}
\end{table}

\begin{table*}[t]
\centering\scriptsize
\begin{tabular}{ccccccccccc}
\toprule
Student & Baseline & SKD \shortcite{hinton2015distilling} & FitNets \shortcite{romero2014fitnets} & AT \shortcite{zagoruyko2016paying} & Jacobian \shortcite{srinivas2018knowledge} & FT \shortcite{kim2018paraphrasing} & AB \shortcite{heo2019knowledge} & Overhaul \shortcite{heo2019comprehensive} & MSE \\
\midrule\midrule
WRN-16-2 & 72.68 & 73.53 & 73.70 & 73.44 & 73.29 & 74.09 & 73.98 & \textbf{75.59} & 75.54 \\
WRN-16-4 & 77.28 & 78.31 & 78.15 & 77.93 & 77.82 & 78.28 & 78.64 & 78.20 & \textbf{79.03} \\
WRN-28-2 & 75.12 & 76.57 & 76.06 & 76.20 & 76.30 & 76.59 & 76.81 & 76.71 & \textbf{77.28} \\
\bottomrule
\end{tabular}
\caption{\textbf{Test accuracy of various KD methods} on CIFAR-100. All student models share the same teacher model as WRN-28-4. The standard KD (SKD) represents the KD method \protect\cite{hinton2015distilling} with hyperparameter values ($\alpha$, $\tau$) used in Eq.~(\ref{eq:KD_loss}) as (0.1, 5). MSE represents the KD with $\mathcal{L}_{MSE}$ between logits; the Overhaul \protect\cite{heo2019comprehensive} model is reproduced by using our pretrained teacher, and the others are the results reported in \protect\cite{heo2019comprehensive}. The baseline indicates the model trained with $\mathcal{L}_{CE}$ without the teacher model.}
\label{tab:acc_comparison}
\end{table*}

We empirically compared the objectives $\mathcal{L}_{KL}$ and $\mathcal{L}_{MSE}$ in terms of performance gains and measured the distance between the logit distributions. Following the previous analysis, we also focused on ``soft'' targets in $\mathcal{L}^\prime$. \autoref{tab:mse_loss} presents the top-1 test accuracies on CIFAR-100 according to the student learning scheme for various teacher-student pairs. The students trained with $\mathcal{L}_{CE}$ are vanilla models without a teacher. The students trained with $\mathcal{L}_{KL}$ or $\mathcal{L}_{MSE}$ are trained following the KD framework without using the ``hard'' targets, i.e., $\alpha=1$ in $\mathcal{L}$ and $\mathcal{L}^\prime$, respectively. It is shown that distillation with $\mathcal{L}_{MSE}$, that is, direct logit distillation without hindering term $\delta_\infty$, is the best training scheme for various teacher-student pairs. We also found the consistent improvements in ensemble distillation\,\cite{hinton2015distilling}. For the ensemble distillation using MSE loss, an ensemble of logit predictions\,(i.e., an average of logit predictions) are used by multiple teachers. We obtained the test accuracy of WRN16-2\,(\textbf{$75.60\%$}) when the WRN16-4, WRN-28-4, and WRN-40-6 models were used as ensemble teachers in this manner. Moreover, the model trained with $\mathcal{L}_{MSE}$ has similar or better performance when compared to other existing KD methods, as described in \autoref{tab:acc_comparison}.\footnote{We excluded the additional experiments for the replacement with MSE loss in feature-based distillation methods. It is difficult to add the MSE loss or replace the KL loss with MSE loss in the existing works because of the sensitivity to hyperparameter optimization. Their methods included various types of hyperparameters that need to be optimized for their settings.}

\begin{figure}[t]
    \centering
    \begin{subfigure}[b]{0.98\textwidth}  
        \centering 
        \includegraphics[width=\textwidth]{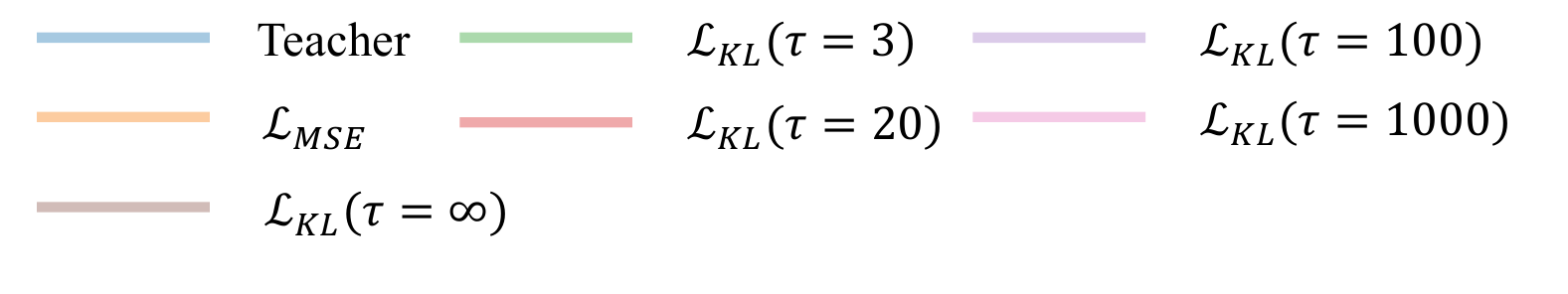}
    \vspace{-20pt}
    \end{subfigure}
    \centering
    \begin{subfigure}[b]{0.48\textwidth}   
        \centering 
        \includegraphics[width=\textwidth]{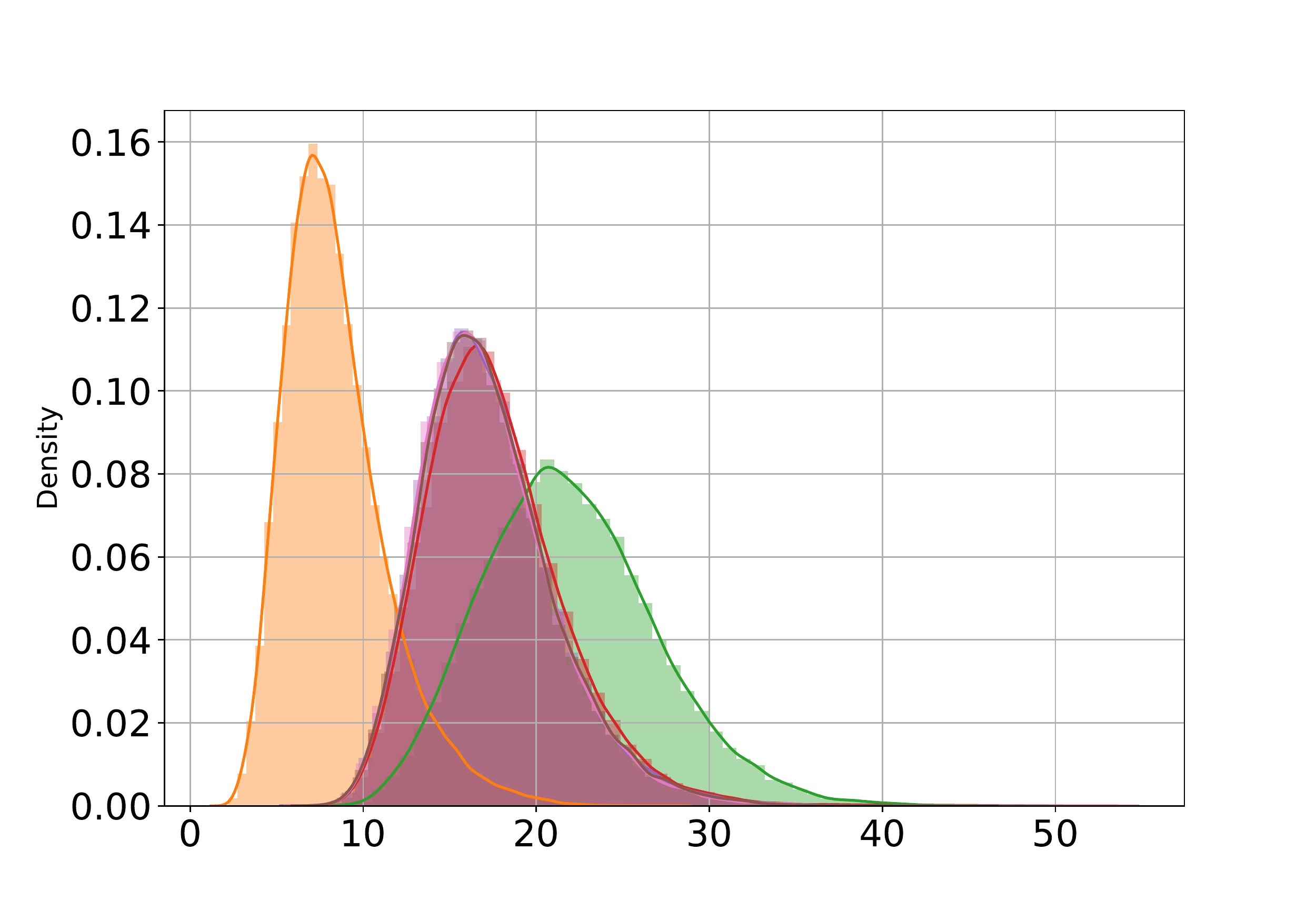}
        \caption{$\| \vz^s - \vz^t \|_2$}
        \label{fig:separate_loss}
    \end{subfigure}
    \begin{subfigure}[b]{0.48\textwidth}  
        \centering 
        \includegraphics[width=\textwidth]{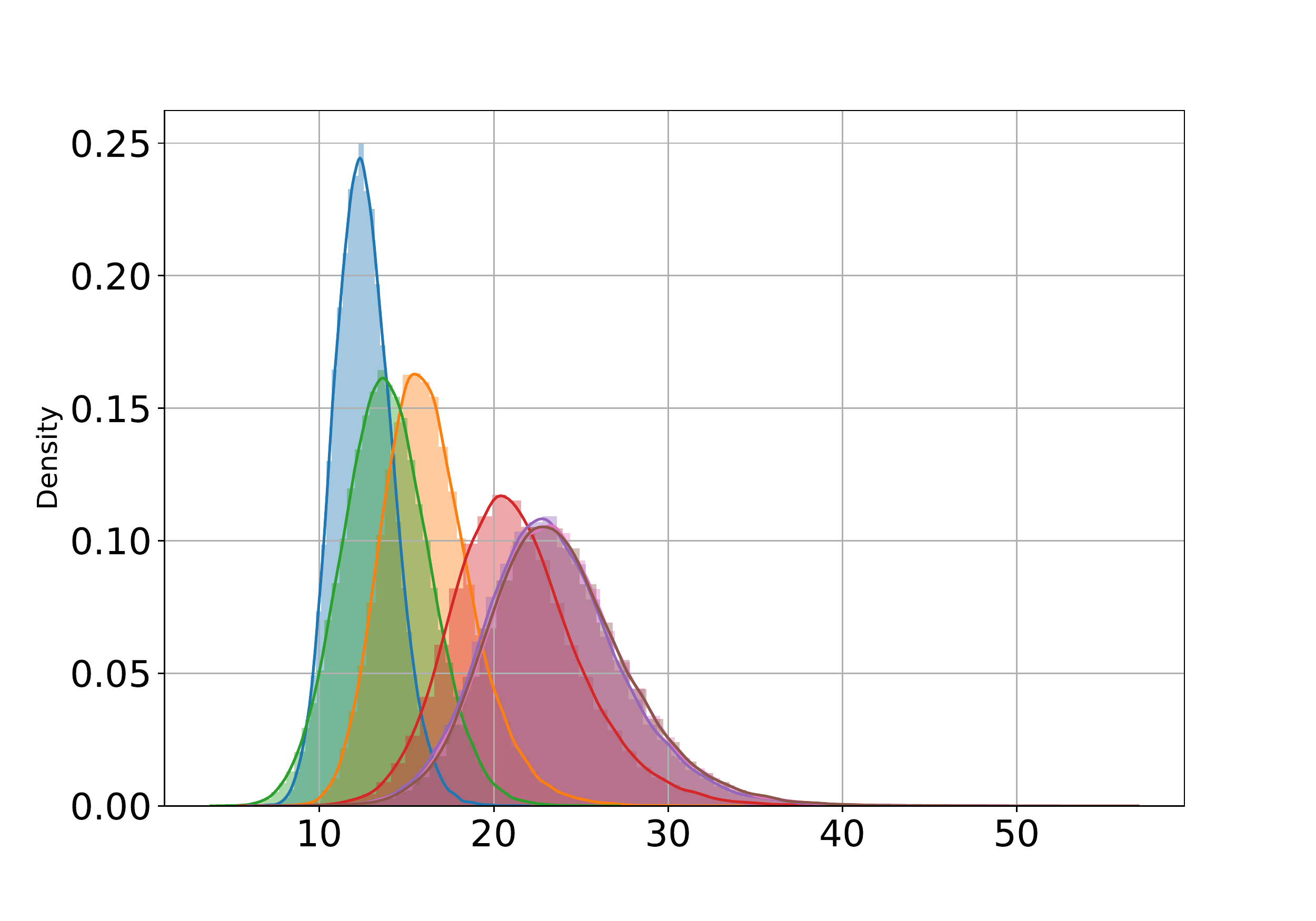}
        \caption{Norm of pre-logit}
        \label{fig5:penultimate_norm}
    \end{subfigure}
    \centering
    \caption{\textbf{(a)} Probabilistic density function\,(pdf) for $\| \vz^s - \vz^t \|_2$ on CIFAR-100 training dataset; \textbf{(b)} The pdf for the 2-norm of pre-logit (i.e., $\| r^s \|_2$) on CIFAR-100 training dataset. We use a (teacher, student) pair of (WRN-28-4, WRN-16-2).}
    \label{fig:norm}
\end{figure}

Furthermore, to measure the distance between the student's logit $\vz^s$ and the teacher's logit $\vz^t$ sample by sample, we describe the probabilistic density function (pdf) from the histogram for $\| \vz^s - \vz^t \|_2$ on the CIFAR-100 training dataset (\autoref{fig:norm}(a)). The logit distribution of the student with a large $\tau$ is closer to that of the teacher than with a small $\tau$ when $\mathcal{L}_{KL}$ is used. Moreover, $\mathcal{L}_{MSE}$ is more efficient in transferring the teacher's information to a student than $\mathcal{L}_{KL}$. Optimizing $\mathcal{L}_{MSE}$ aligns the student's logit with the teacher's logit. On the other hand, when $\tau$ becomes significantly large, $\mathcal{L}_{KL}$ has the $\delta_{\infty}$, and optimizing $\delta_\infty$ makes the student's logit mean deviate from that of the teacher's logit mean. 

We further investigate the effect of $\delta_\infty$ on the penultimate layer representations\, (i.e., pre-logits). Based on $\delta_\infty \approx  - \frac{1}{2K^2} ({\sum_{j=1}^K \vz_j^s})^2$, we can reformulate Eq.~(\ref{eq:relation}). Let $\vr^s \in \mathbb{R}^{d}$ be the penultimate representation of student $s$ from an instance $\vx$, and $W^s \in \mathbb{R}^{K \times d}$ be the weight matrix of the student's fully connected layer. Then,

\begin{equation}\label{eq:shrinkage}
\begin{split}
    \delta_\infty & \approx  - \frac{1}{2K^2} \left( {\sum_{j=1}^K \vz_j^s} \right)^2 = - \frac{1}{2K^2} \left( \sum_{j=1}^K\sum_{n=1}^{d} W^s_{j,n}\vr^s_n \right)^2 \\
    & = - \frac{1}{2K^2} \left( \sum_{n=1}^{d} \vr^s_n \sum_{j=1}^K W^s_{j,n} \right)^2 \\
    & \geq - \frac{1}{2K^2} \left( \sum_{n=1}^{d} (\sum_{j=1}^K W^s_{j,n})^2 \right) \left(\sum_{n=1}^{d} {\vr^s_n}^2 \right) \\ 
    & \quad (\because \text{Cauchy-Schwartz inequality})\\
    & = - \frac{1}{2K^2} ||\vr^s||_2^2 \left( \sum_{n=1}^{d} (\sum_{j=1}^K W^s_{j,n})^2 \right) \\ 
\end{split}
\end{equation}

\begin{figure}[h]
    \centering
    \begin{subfigure}[b]{0.48\textwidth}  
        \centering 
        \includegraphics[width=\textwidth]{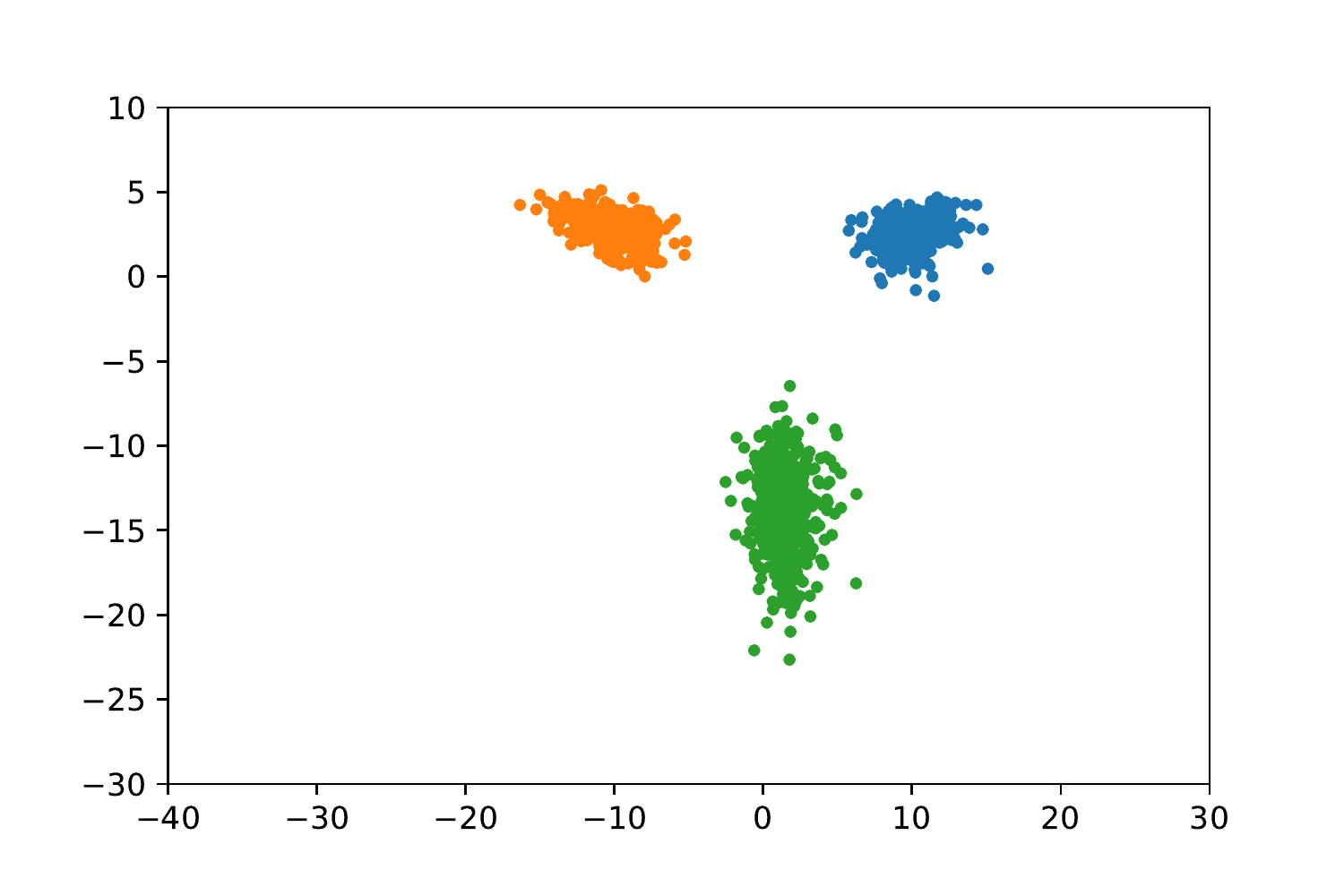}
        \caption{$t, \mathcal{L}_{CE}$ (Train)}
        \label{fig4:a}
    \end{subfigure}
    \begin{subfigure}[b]{0.48\textwidth}  
        \centering 
        \includegraphics[width=\textwidth]{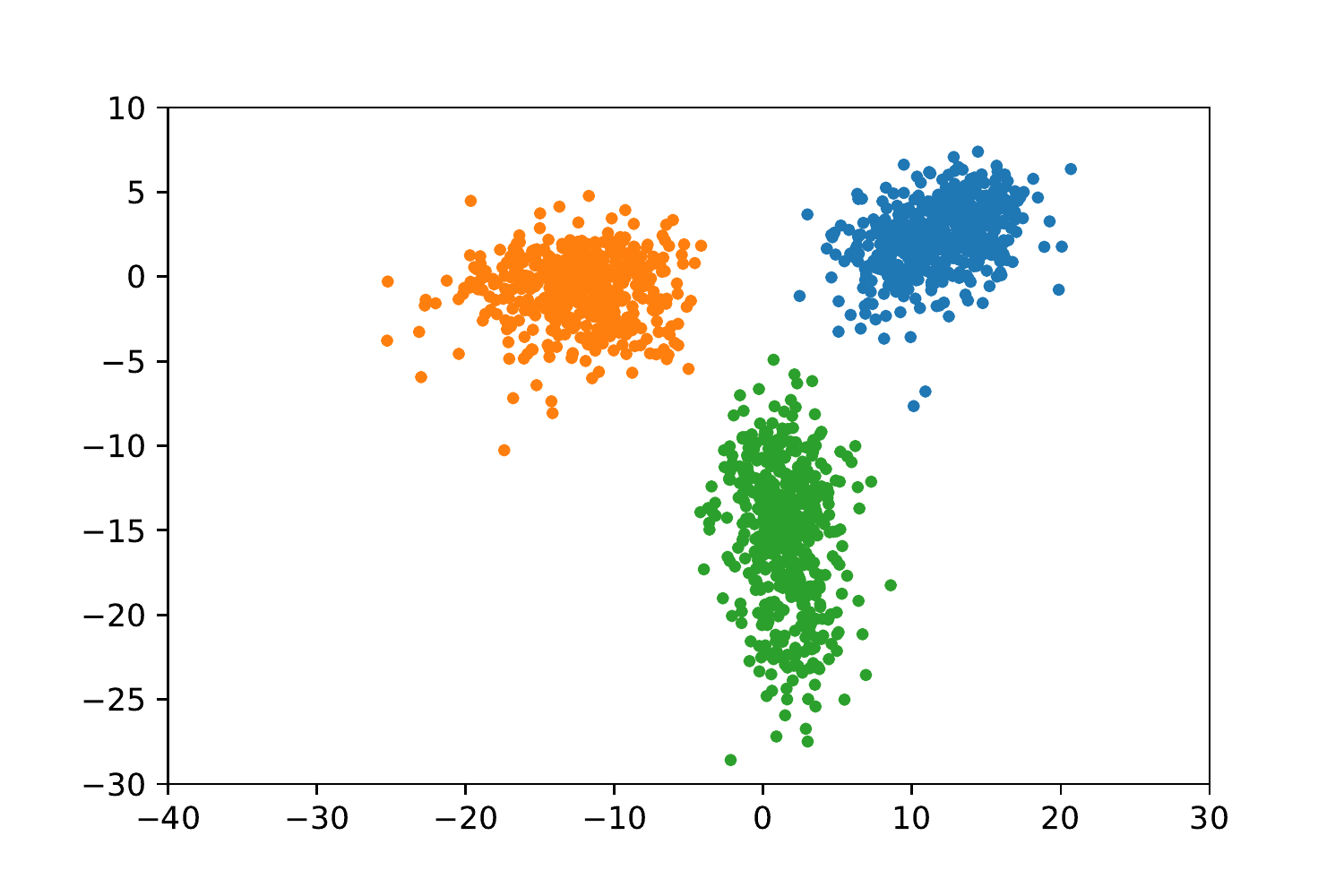}
        \caption{$s, \mathcal{L}_{CE}$ (Train)}
        \label{fig4:e}
    \end{subfigure}
    \begin{subfigure}[b]{0.48\textwidth}  
        \centering 
        \includegraphics[width=\textwidth]{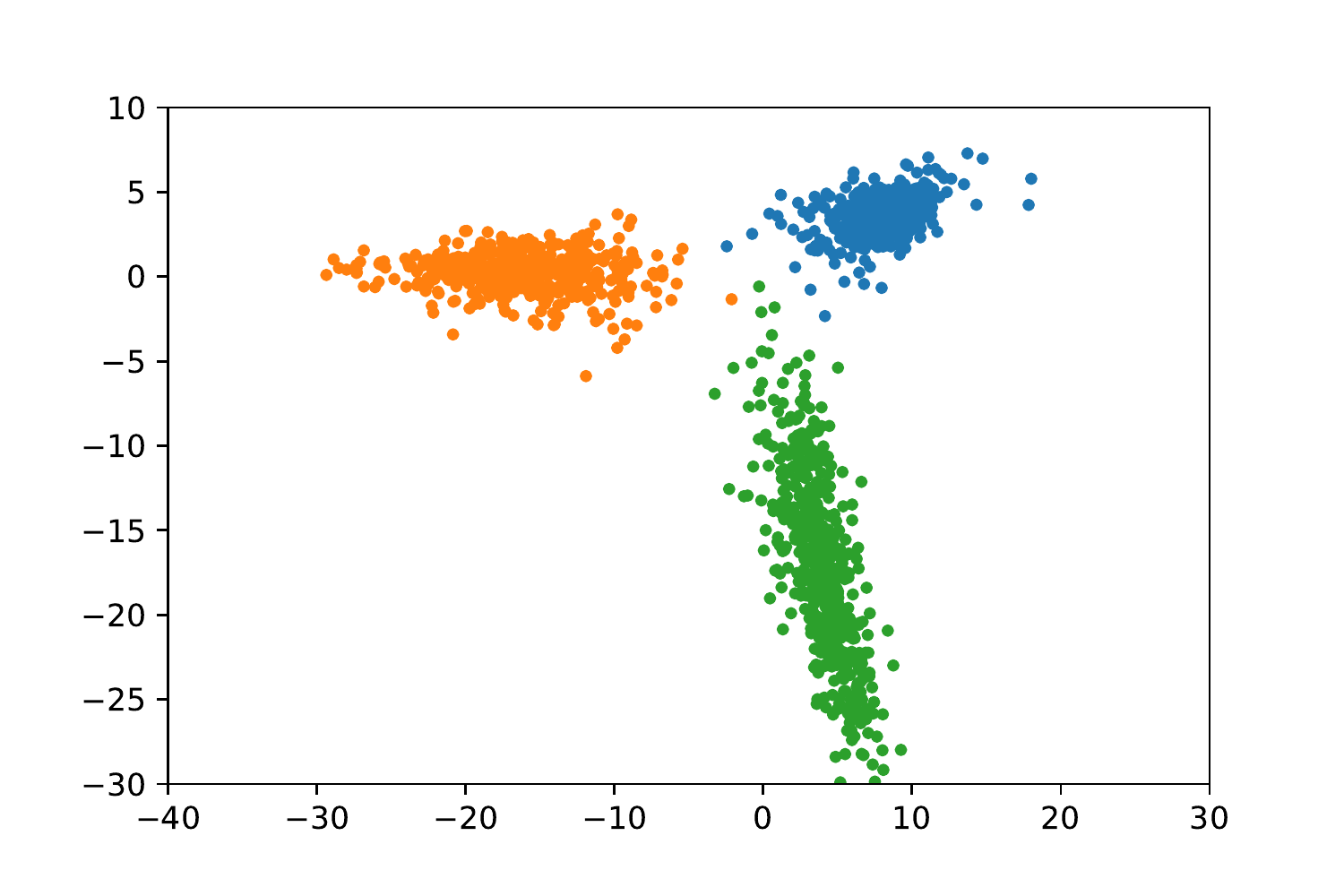}
        \caption{$s, \mathcal{L}_{KL}(\tau=\infty)$ (Train)}
        \label{fig4:i}
    \end{subfigure}
    \begin{subfigure}[b]{0.48\textwidth}  
        \centering 
        \includegraphics[width=\textwidth]{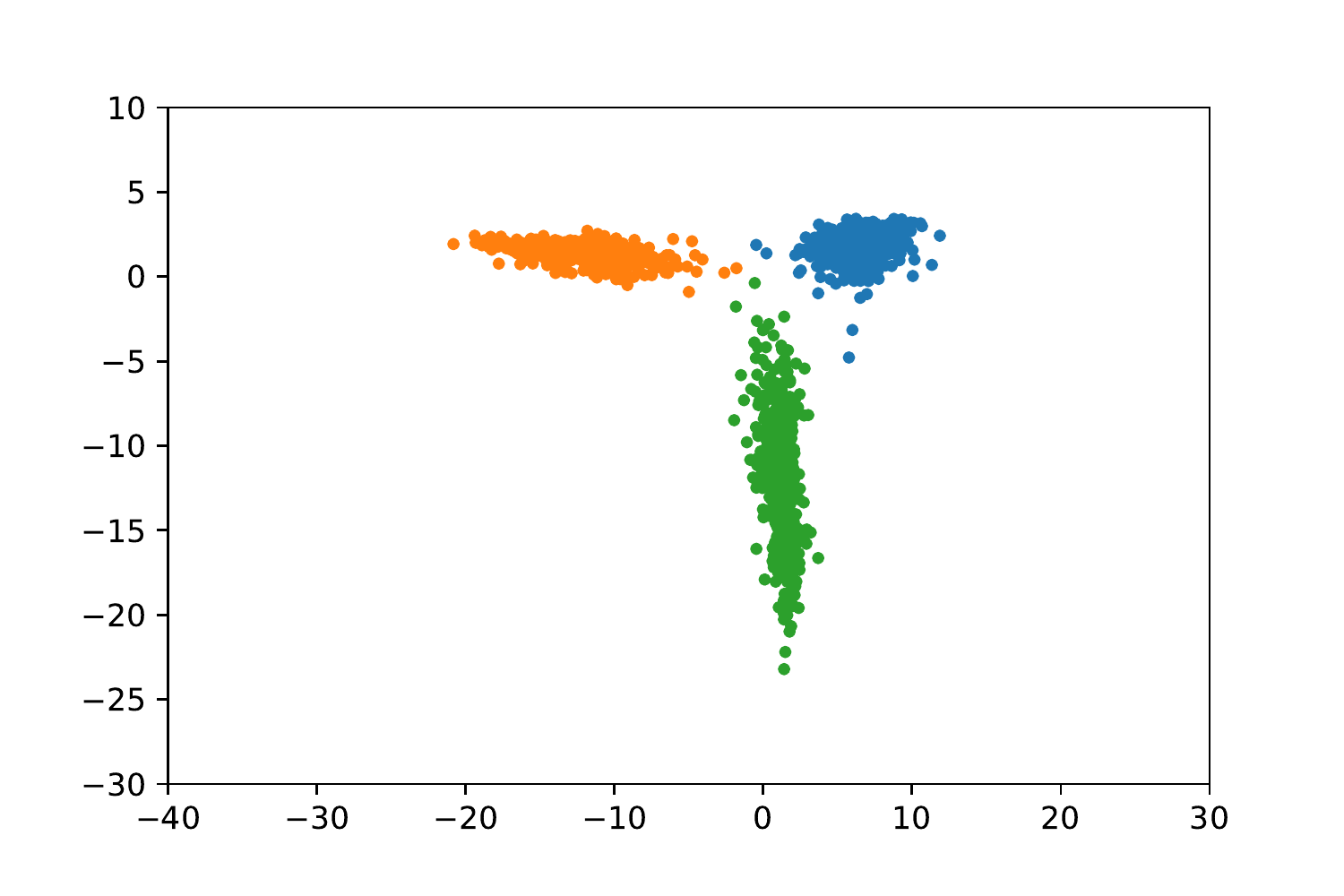}
        \caption{$s, \mathcal{L}_{MSE}$ (Train)}
        \label{fig4:m}
    \end{subfigure}
    \centering
    \caption{Visualizations of pre-logits on CIFAR-100 according to the change of loss function. Here, we use the classes ``apple,'' ``aquarium fish,'' and ``baby.'' $t$ indicates the teacher network (WRN-28-4), and $s$ indicates the student network (WRN-16-2).}
    \label{fig:ls_visualization}
\end{figure}

As derived in Eq.~(\ref{eq:shrinkage}), training a network with $\mathcal{L}_{KL}$ encourages the pre-logits to be dilated via $\delta_\infty$\,(\autoref{fig5:penultimate_norm}). For visualization, following \cite{muller2019does}, we first find an orthonormal basis constructed from the templates (i.e., the mean of the representations of the samples within the same class) of the three selected classes (apple, aquarium fish, and baby in our experiments). Then, the penultimate layer representations are projected onto the hyperplane based on the identified orthonormal basis. WRN-28-4 ($t$) is used as a teacher, and WRN-16-2 ($s$) is used as a student on the CIFAR-100 training dataset. As shown in the first row of \autoref{fig:ls_visualization}, when WRN-like architectures are trained with $\mathcal{L}_{CE}$ based on ground-truth hard targets, clusters are tightened as the model's complexity increases. As shown in the second row of \autoref{fig:ls_visualization}, when the student $s$ is trained with $\mathcal{L}_{KL}$ with infinite $\tau$ or with $\mathcal{L}_{MSE}$, both representations attempt to follow the shape of the teacher's representations but differ in the degree of cohesion. This is because $\delta_\infty$ makes the pre-logits become much more widely clustered. Therefore, $\mathcal{L}_{MSE}$ can shrink the representations more than $\mathcal{L}_{KL}$ along with the teacher.

\subsection{Effects of a Noisy Teacher}
We investigate the effects of a noisy teacher\,(i.e., a model poorly fitted to the training dataset) according to the objective. It is believed that the \emph{label matching} ($\mathcal{L}_{KL}$ with a small $\tau$) is more appropriate than the \emph{logit matching} ($\mathcal{L}_{KL}$ with a large $\tau$ or the $\mathcal{L}_{MSE}$) under a noisy teacher. This is because \emph{label matching} neglects the negative information of the outputs of an untrained teacher. \autoref{tab:untrained} describes top-1 test accuracies on CIFAR-100, where the used teacher network (WRN-28-4) has a training accuracy of 53.77\%, which is achieved in 10 epochs. When poor knowledge is distilled, the students following the \emph{label matching} scheme performed better than the students following the \emph{logit matching} scheme, and the extreme \emph{logit matching} through $\mathcal{L}_{MSE}$ has the worst performance. Similarly, it seems that \emph{logit matching} is not suitable for large-scale tasks. \autoref{tab:ImageNet} presents top-1 test accuracies on ImageNet, where the used teacher network (ResNet-152) has a training accuracy of 81.16\%, which is provided in PyTorch. Even in this case, the extreme \emph{logit matching} exhibits the worst performance. The utility of negative logits (i.e., negligible aspect when $\tau$ is small) was discussed in \cite{hinton2015distilling}.


\section{Sequential Distillation}
In \cite{cho2019efficacy}, the authors showed that more extensive teachers do not mean better teachers, insisting that the capacity gap between the teacher and the student is a more important factor than the teacher itself. In their results, using a medium-sized network instead of a large-scale network as a teacher can improve the performance of a small network by reducing the capacity gap between the teacher and the student. They also showed that sequential KD (large network $\rightarrow$ medium network $\rightarrow$ small network) is not conducive to generalization when $(\alpha, \tau)=(0.1, 4)$ in Eq.~(\ref{eq:KD_loss}). In other words, the best approach is a direct distillation from the medium model to the small model.

\autoref{tab:sequent} describes the test accuracies of sequential KD, where the largest model is WRN-28-4, the intermediate model is WRN-16-4, and the smallest model is WRN-16-2. Similar to the previous study, when $\mathcal{L}_{KL}$ with $\tau=3$ is used to train the small network iteratively, the direct distillation from the intermediate network to the small network is better (i.e., WRN-16-4 $\rightarrow$ WRN-16-2, 74.84\%) than the sequential distillation (i.e., WRN-28-4 $\rightarrow$ WRN-16-4 $\rightarrow$ WRN-16-2, 74.52\%) and direct distillation from a large network to a small network (i.e., WRN-28-4 $\rightarrow$ WRN-16-2, 74.24\%). The same trend occurs in $\mathcal{L}_{MSE}$ iterations.

On the other hand, we find that the medium-sized teacher can improve the performance of a smaller-scale student when $\mathcal{L}_{KL}$ and $\mathcal{L}_{MSE}$ are used sequentially (the last fourth row) despite the large capacity gap between the teacher and the student. KD iterations with such a strategy might compress the model size more effectively, and hence should also be considered in future work. Furthermore, our work is the first study on the sequential distillation at the \emph{objective} level, not at the \emph{architecture} level such as \cite{cho2019efficacy,mirzadeh2020improved}.

\begin{table}[t]
\centering
\scriptsize
\begin{tabular}{c|ccccccc}
    \toprule
    \multirow{2}{*}{Student} & \multicolumn{6}{c}{$\mathcal{L}_{KL}$} & \multirow{2}{*}{$\mathcal{L}_{MSE}$} \\
    \cmidrule{2-7}
     & $\tau$=0.1 & $\tau$=0.5 & $\tau$=1 & $\tau$=5 & $\tau$=20 & $\tau$=$\infty$ &  \\
    \midrule\midrule
    WRN-16-2 & 51.64 & 52.07 & 51.36 & 50.11 & 49.69 & 49.46 & 49.20 \\
    \bottomrule
\end{tabular}
\caption{Top-1 test accuracies on CIFAR-100. WRN-28-4 is used as a teacher for $\mathcal{L}_{KL}$ and $\mathcal{L}_{MSE}$. Here, the teacher (WRN-28-4) was not fully trained. The training accuracy of the teacher network is 53.77\%.}
\label{tab:untrained}
\end{table}

\begin{table}[t]
\centering
\small
\begin{tabular}{c|cccc}
    \toprule
    Student & $\mathcal{L}_{CE}$ & $\mathcal{L}_{KL}$ (Standard) & $\mathcal{L}_{KL}$ ($\tau$=20) & $\mathcal{L}_{MSE}$ \\
    \midrule\midrule
    ResNet-50 & 76.28 & 77.15 & 77.52 & 75.84 \\
    \bottomrule
    \end{tabular}
    \caption{Test accuracy on the ImageNet dataset. We used a (teacher, student) pair of (ResNet-152, ResNet-50). We include the results of the baseline and $\mathcal{L}_{KL}$ (standard) from \protect\cite{heo2019comprehensive}. The training accuracy of the teacher network is 81.16\%.}
    \label{tab:ImageNet}
\end{table}

\begin{table}[t]
\centering
\footnotesize\addtolength{\tabcolsep}{-2.5pt}
\begin{tabular}{cccc}
    \toprule
    WRN-28-4 & WRN-16-4 & WRN-16-2 & Test accuracy \\
    \midrule \midrule 
    X & X & $\mathcal{L}_{CE}$ & 72.68 \% \\
    \midrule
    \multirow{4}{*}{X} & \multirow{4}{*}{ \makecell{$\mathcal{L}_{CE}$\\(77.28\%)}} & { $\mathcal{L}_{KL}$($\tau=3$)} & 74.84 \% \\ \cmidrule{3-4}
     &  & {$\mathcal{L}_{KL}$($\tau=20$)} & 75.42 \% \\ \cmidrule{3-4}
     &  & {$\mathcal{L}_{MSE}$} & 75.58 \%\\ \midrule
     \multirow{4}{*}{\makecell{$\mathcal{L}_{CE}$\\(78.88\%)}} & \multirow{4}{*}{X} & { $\mathcal{L}_{KL}$($\tau=3$)} & 74.24 \% \\ \cmidrule{3-4}
     &  & {$\mathcal{L}_{KL}$($\tau=20$)} & 75.15 \% \\ \cmidrule{3-4}
     &  & {$\mathcal{L}_{MSE}$} & 75.54 \%\\ \midrule
    \multirow{8}{*}{\makecell{$\mathcal{L}_{CE}$\\(78.88\%)}} & \multirow{4}{*}{\makecell{$\mathcal{L}_{KL}$($\tau=3$) \\ (78.76\%)}} & \makecell{$\mathcal{L}_{KL}$($\tau=3$)} & 74.52 \% \\ \cmidrule{3-4}
     &  & \makecell{$\mathcal{L}_{KL}$($\tau=20$)} & 75.47 \% \\ \cmidrule{3-4}
     &  & \makecell{$\mathcal{L}_{MSE}$} & \textbf{75.78 \%} \\ \cmidrule{2-4}
     &  \multirow{4}{*}{\makecell{$\mathcal{L}_{MSE}$ \\ (79.03\%)}} & \makecell{$\mathcal{L}_{KL}$($\tau=3$)} & 74.83 \% \\ \cmidrule{3-4}
     &  & \makecell{$\mathcal{L}_{KL}$($\tau=20$)} & 75.47 \% \\ \cmidrule{3-4}
     &  & \makecell{$\mathcal{L}_{MSE}$} & 75.48 \% \\ \bottomrule
\end{tabular}
\caption{Test accuracies of sequential knowledge distillation. In each entry, we note the objective function that used for the training. `X' indicates that distillation was not used in training.}
\label{tab:sequent}
\end{table}

\section{Robustness to Noisy Labels}
In this section, we investigate how \emph{noisy labels}, samples annotated with incorrect labels in the training dataset, affect the distillation ability when training a teacher network. This setting is related to the capacity for memorization and generalization. Modern deep neural networks even attempt to memorize samples perfectly\,\cite{zhang2016understanding}; hence, the teacher might transfer \emph{corrupted knowledge} to the student in this situation. Therefore, it is thought that logit matching might not be the best strategy when the teacher is trained using a noisy label dataset.



\begin{figure}[t]
    \centering
    \begin{subfigure}[b]{0.98\textwidth}  
        \centering 
        \includegraphics[width=\textwidth]{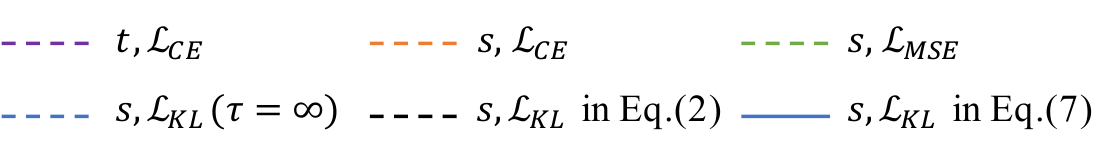}
    \end{subfigure}
    \begin{subfigure}[b]{0.49\textwidth}  
        \centering 
        \includegraphics[width=\textwidth]{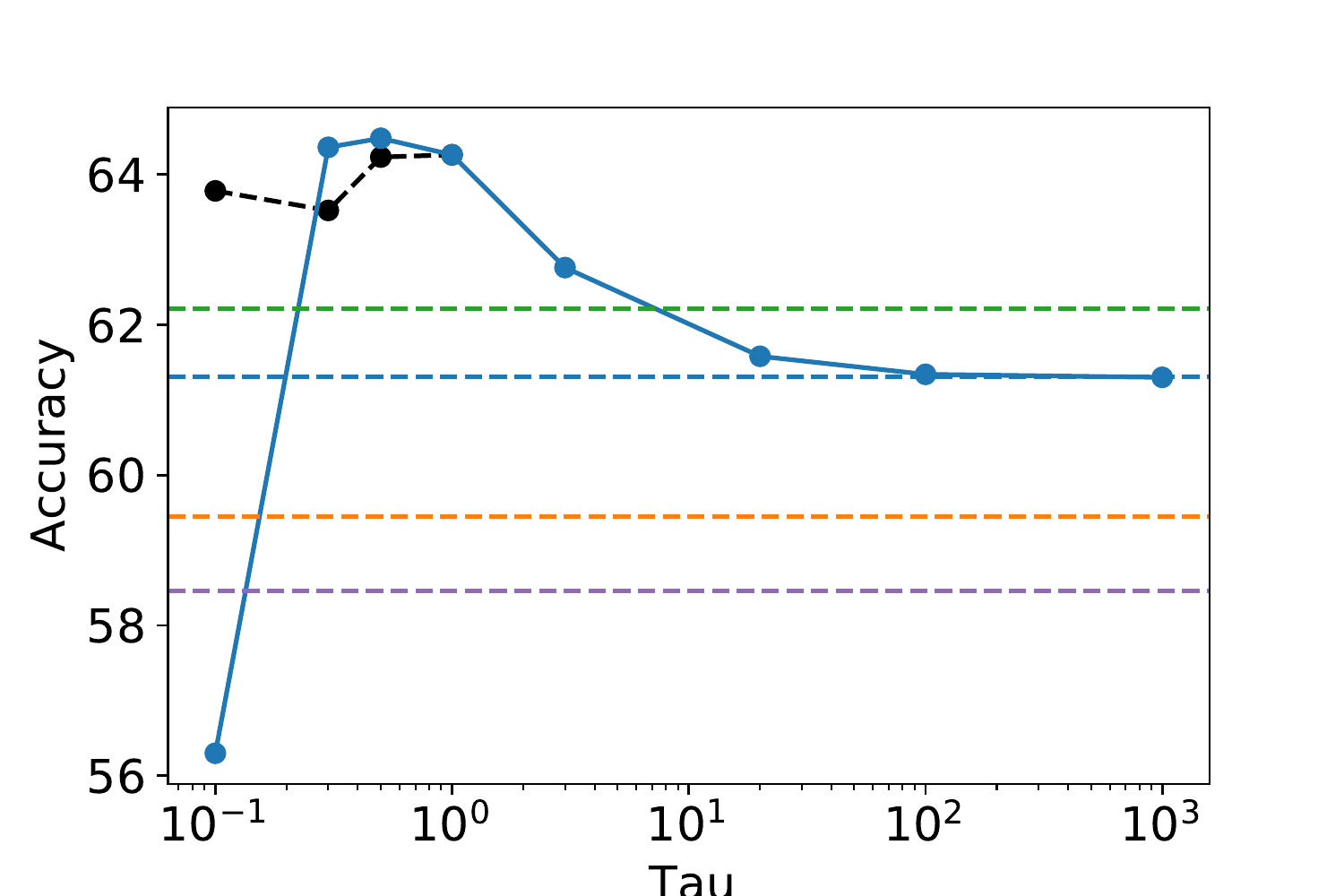}
        \caption{Symmetric noise 40\%}
        \label{fig:sym_40}
    \end{subfigure}
    \begin{subfigure}[b]{0.49\textwidth}   
        \centering 
        \includegraphics[width=\textwidth]{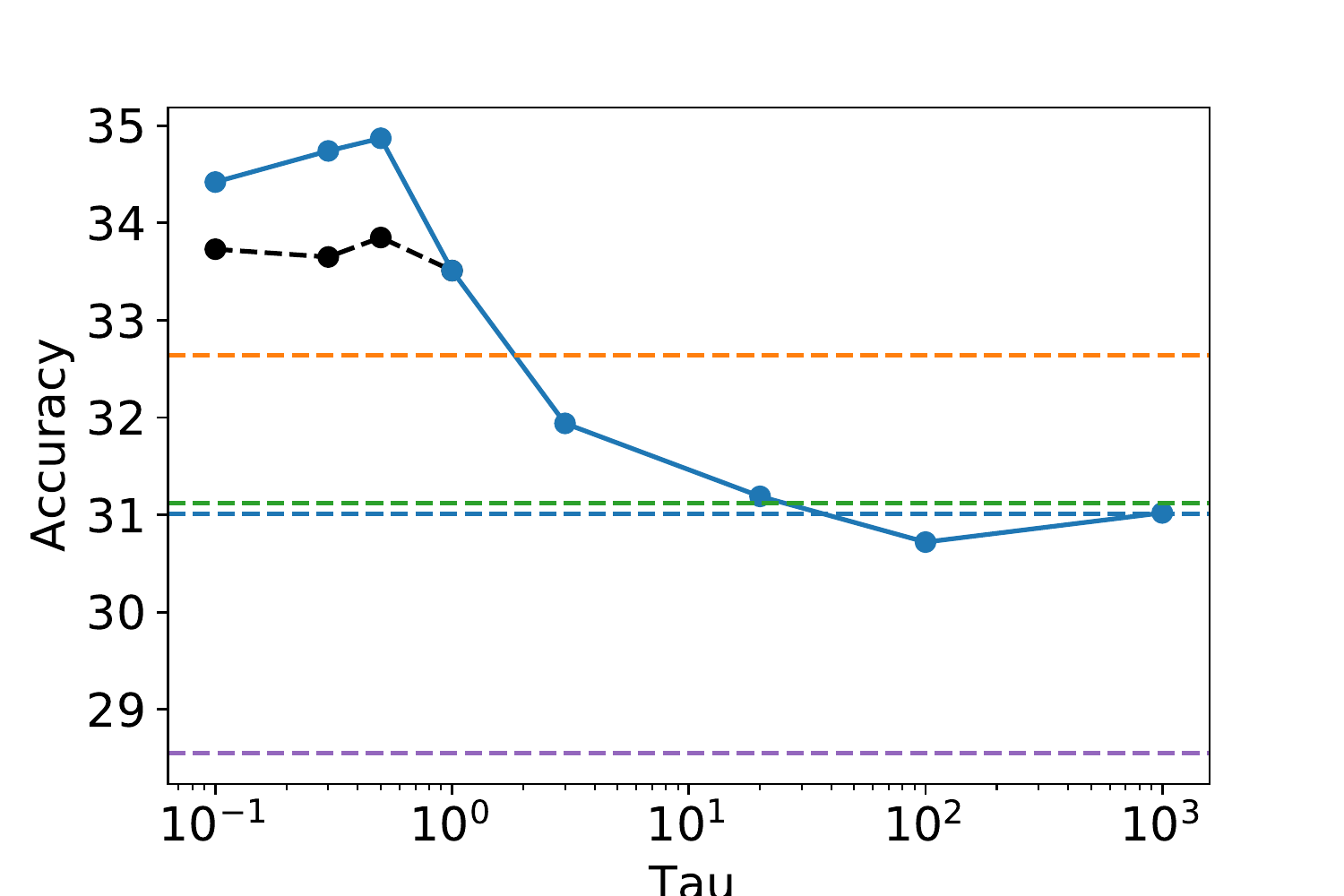}
        \caption{Symmetric noise 80\%}
        \label{fig:sym_80}
    \end{subfigure}
    \centering
    \caption{Test accuracy graph as $\tau$ changes on CIFAR-100. We use the (teacher, student) as (WRN-28-4, WRN-16-2).}
    \label{fig:noise}
\end{figure}

From this insight, we simulate the noisy label setting to evaluate the robustness on CIFAR-100 by randomly flipping a certain fraction of the labels in the training dataset following a symmetric uniform distribution. \autoref{fig:noise} shows the test accuracy graphs as the loss function changes. First, we observe that a small network (WRN-16-2 ($s$), orange dotted line) has a better generalization performance than an extensive network (WRN-28-4 ($t$), purple dotted line) when models are trained with $\mathcal{L}_{CE}$. This implies that a complex model can memorize the training dataset better than a simple model, but cannot generalize to the test dataset. Next, WRN-28-4 (purple dotted line) is used as the teacher model. When the noise is less than 50\%, extreme logit matching ($\mathcal{L}_{MSE}$, green dotted line) and logit matching with $\delta_\infty$ ($\mathcal{L}_{KL}(\tau=\infty)$, blue dotted line) can mitigate the label noise problem compared with the model trained with $\mathcal{L}_{CE}$. However, when the noise is more than 50\%, these training cannot mitigate this problem because it follows corrupted knowledge more often than correct knowledge.

Interestingly, the best generalization performance is achieved when we use $\mathcal{L}_{KL}$ with $\tau \le 1.0$. In \autoref{fig:noise}, the blue solid line represents the test accuracy using the rescaled loss function from the black dotted line when $\tau \le 1.0$. As expected, logit matching might transfer the teacher's overconfidence, even for incorrect predictions. However, the proper objective derived from both logit matching and label matching enables similar effects of label smoothing, as studied in \cite{pmlr-v119-lukasik20a,yuan2020revisit}. Therefore, $\mathcal{L}_{KL}$ with $\tau=0.5$ appears to significantly mitigate the problem of noisy labels. 



\section{Conclusion}
In this paper, we first showed the characteristics of a student trained with $\mathcal{L}_{KL}$ according to the temperature-scaling hyperparameter $\tau$. As $\tau$ goes to 0, the trained student has the \emph{label matching} property. In contrast, as $\tau$ goes to $\infty$, the trained student has the \emph{logit matching} property. Nevertheless, $\mathcal{L}_{KL}$ with a sufficiently large $\tau$ cannot achieve complete logit matching owing to $\delta_\infty$. To achieve this goal, we proposed a direct logit learning framework using $\mathcal{L}_{MSE}$ and improved the performance based on this loss function. In addition, we showed that the model trained with $\mathcal{L}_{MSE}$ followed the teacher's penultimate layer representations more than that with $\mathcal{L}_{KL}$. We observed that sequential distillation can be a better strategy when the capacity gap between the teacher and the student is large. Furthermore, we empirically observed that, in the noisy label setting, using $\mathcal{L}_{KL}$ with $\tau$ near 1 mitigates the performance degradation rather than extreme logit matching, such as $\mathcal{L}_{KL}$ with $\tau=\infty$ or $\mathcal{L}_{MSE}$.
\section*{Acknowledgments}
This work was supported by Institute of Information \& communications Technology Planning \& Evaluation (IITP) grant funded by the Korea government (MSIT) [No.2019-0-00075, Artificial Intelligence Graduate School Program (KAIST)] and [No. 2021-0-00907, Development of Adaptive and Lightweight Edge-Collaborative Analysis Technology for Enabling Proactively Immediate Response and Rapid Learning].

\bibliographystyle{named}
\bibliography{ijcai21}

\clearpage

\appendix
\onecolumn

\section{Details of formulas}\label{sec:detail_sec2}
\subsection{Gradient of KD loss}

\begin{equation}
    \begin{aligned}
    \mathcal{L} & = (1-\alpha) \mathcal{L}_{CE}(\vp^{s}(1), \vy) + \alpha \mathcal{L}_{KL}(\vp^{s}(\tau), \vp^{t}(\tau)) \\
    & = (1-\alpha) \left[ \sum_{j} - \vy_j \log \vp^{s}_j(1) \right] + \alpha \tau^{2} \left[ \sum_{j} \vp^{t}_j(\tau) \log \vp^{t}_j(\tau) - \vp^{t}_j(\tau) \log \vp^{s}_j(\tau) \right] \\
    \frac{\partial \mathcal{L}}{\partial{\vz_k^s}} & = (1-\alpha) \left[ \sum_{j} - \vy_j \frac{\partial \log \vp^{s}_{j}(1)}{\partial \vz_k^s} \right] + \alpha \tau^{2} \left[ \sum_{j} - \vp^{t}_{j}(\tau) \frac{\partial \log \vp^{s}_{j}(\tau)}{\partial \vz_k^s} \right] \\
    & = (1-\alpha) \left[ \sum_{j} - \frac{\vy_j}{\vp^{s}_{j}(1)} \frac{\partial \vp^{s}_{j}(1)}{\partial \vz_k^s} \right] + \alpha \tau^{2} \left[ \sum_{j} - \frac{\vp^{t}_{j}(\tau)}{\vp^{s}_{j}(\tau)} \frac{\partial \vp^{s}_{j}(\tau)}{\partial \vz_k^s} \right] \, \text{where} \, \vp^{f}_{j}(\tau) = \frac{ e^{\vz_j^f/\tau} }{ \sum_{i} e^{\vz_i^f/\tau} } \\
    & = (1-\alpha) \left[ - \frac{\vy_{k}}{\vp^{s}_{k}(1)} \frac{\partial \vp^{s}_{k}(1)}{\partial \vz_k^s} \right] + \alpha \tau^{2} \left[ - \frac{\vp^{t}_{k}(\tau)}{\vp^{s}_{k}(\tau)} \frac{\partial \vp^{s}_{k}(\tau)}{\partial \vz_k^s} + \sum_{j \neq k} -  \frac{\vp^{t}_{j}(\tau)}{\vp^{s}_{k}(\tau)} \frac{\partial \vp^{s}_{j}(\tau)}{\partial \vz_k^s} \right] \\
    & \hspace{280pt} (\because \vy_j=0 \, \text{for all} \, j \neq k) \\
    & = (1-\alpha) (\vp^{s}_{k}(1) - \vy_k) + \alpha \left[ \tau (\vp^{s}_{k}(\tau) - \vp^{t}_{k}(\tau)) \right] \\
    & \hspace{75pt} \left( \because \frac{\partial \vp^{s}_{j}(\tau)}{\partial \vz_k^s} = \partial \left( \frac{ e^{\vz_j^s/\tau} }{ \sum_{i} e^{\vz_i^s/\tau} } \right) / \partial \vz_k^s = \begin{cases} \frac{1}{\tau} \vp^{s}_{k}(\tau) (1-\vp^{s}_{k}(\tau)), \; \text{if} \, j=k \\ -\frac{1}{\tau} \vp^{s}_{j}(\tau)\vp^{s}_{k}(\tau), \; \; \; \; \; \text{otherwise} \end{cases} \hspace{-5pt} \right)
    \end{aligned}
\label{eq:gradient_proof}
\end{equation}

\subsection{Proof of Proposition \ref{thm1}}
\begin{equation}
    \begin{aligned}
    \lim_{\tau \to \infty} \tau (\vp^{s}_{k}(\tau) - \vp^{t}_{k}(\tau)) & = \lim_{\tau \to \infty} \tau \left( \frac{e^{\vz^s_k/\tau}}{\sum_{j=1}^K e^{\vz^s_j/\tau}} - \frac{e^{\vz^t_k/\tau}}{\sum_{j=1}^K e^{\vz^t_j/\tau}} \right) \\
    & = \lim_{\tau \to \infty} \tau \left( \frac{1}{\sum_{j=1}^K e^{(\vz^s_j-\vz^s_k)/\tau}} - \frac{1}{\sum_{j=1}^K e^{(\vz^t_j-\vz^t_k)/\tau}} \right) \\
    & = \lim_{\tau \to \infty} \tau \left( \frac{\sum_{j=1}^K e^{(\vz^t_j-\vz^t_k)/\tau} - \sum_{j=1}^K e^{(\vz^s_j-\vz^s_k)/\tau}}{ \left( \sum_{j=1}^K e^{(\vz^s_j-\vz^s_k)/\tau} \right) \left( {\sum_{j=1}^K e^{(\vz^t_j-\vz^t_k)/\tau}} \right)} \right) \\
    & = \lim_{\tau \to \infty} \left( \frac{ \sum_{j=1}^K \tau \left( e^{(\vz^t_j-\vz^t_k)/\tau} - 1 \right) - \sum_{j=1}^K \tau \left( e^{(\vz^s_j-\vz^s_k)/\tau} - 1 \right) }{ \left( \sum_{j=1}^K e^{(\vz^s_j-\vz^s_k)/\tau} \right) \left( {\sum_{j=1}^K e^{(\vz^t_j-\vz^t_k)/\tau}} \right)} \right) \\
    & = \frac{1}{K^2} \sum_{j=1}^K \left( (\vz^s_k - \vz^s_j) -(\vz^t_k - \vz^t_j) \right) = \frac{1}{K} \left( \vz^s_k - \vz^t_k \right ) - \frac{1}{K^2} \sum_{j=1}^K \left( \vz^s_j - \vz^t_j \right ) \\
    & \hspace{35pt} \left( \because \lim_{\tau \to \infty} e^{ (\vz^f_j-\vz^f_k) / {\tau}} = 1, \lim_{\tau \to \infty} \tau \left( e^{ (\vz^f_j-\vz^f_k) / {\tau}} -1 \right) = \vz^f_j-\vz^f_k \right) \\
    \end{aligned}
\label{eq:lim_to_inf}
\end{equation}

\begin{equation}
    \begin{aligned}
    \lim_{\tau \to 0} \tau (\vp^{s}_{k}(\tau) - \vp^{t}_{k}(\tau)) = 0 \hspace{5pt} (\because -1 \leq  \vp^{s}_{k}(\tau) - \vp^{t}_{k}(\tau) \leq 1) \hspace{145pt}
    \end{aligned}
\label{eq:lim_to_zero}
\end{equation}


\subsection{Proof of \autoref{eq:relation}}
To prove Eq.~(\ref{eq:relation}), we use the bounded convergence theorem (BCT) to interchange of limit and integral. Namely, it is sufficient to prove that $\lim_{\tau \to \infty} |g_k(\tau)|$ is bounded, where $g_k(\tau) = \frac{\partial \mathcal{L}_{KL}}{\partial{\vz_k^s}} = \tau (\vp^{s}_{k}(\tau) - \vp^{t}_{k}(\tau))$ is each partial derivative. $\forall \tau \in (0, \infty)$,

\begin{equation}\label{eq:bound}
\begin{aligned}
    \lim_{\tau \to \infty} |g_k(\tau)| & = \lim_{\tau \to \infty} \left| \frac{e^{\vz_k^s/\tau} / \sum_{i} e^{\vz_i^s/\tau} - \1/K}{1/\tau}  - \frac{e^{\vz_k^t/\tau} / \sum_{i} e^{\vz_i^t/\tau} - \1/K}{1/\tau} \right| \\
    & \leq \lim_{\tau \to \infty} \left| \frac{e^{\vz_k^s/\tau} / \sum_{i} e^{\vz_i^s/\tau} - \1/K}{1/\tau}  \right| + \lim_{\tau \to \infty} \left| \frac{e^{\vz_k^t/\tau} / \sum_{i} e^{\vz_i^t/\tau} - \1/K}{1/\tau} \right| \\
    & = \lim_{h \to 0} \left| \frac{e^{\vz_k^s \cdot h} / \sum_{i} e^{\vz_i^s \cdot h} - \1/K}{h}  \right| +\lim_{h \to 0} \left| \frac{e^{\vz_k^t\cdot h} / \sum_{i} e^{\vz_i^t\cdot h} - \1/K}{h} \right| \\
    & = \left| \frac{z_k^s e^{\vz_k^s \cdot h} (\sum_{i} e^{\vz_i^s \cdot h}) - e^{\vz_k^s \cdot h} (\sum_{i} \vz_i^s e^{\vz_i^s \cdot h}) }{(\sum_{i} e^{\vz_i^s \cdot h})^2} \right|_{h=0} + \left| \frac{z_k^t e^{\vz_k^t \cdot h} (\sum_{i} e^{\vz_i^t \cdot h}) - e^{\vz_k^t \cdot h} (\sum_{i} \vz_i^t e^{\vz_i^t \cdot h}) }{(\sum_{i} e^{\vz_i^t \cdot h})^2} \right|_{h=0} \\
    & = \left|\frac{Kz_k^s - (\sum_{i} \vz_i^s)}{K^2}\right| + \left|\frac{Kz_k^t - (\sum_{i} \vz_i^t)}{K^2}\right| \quad\quad \blacksquare
\end{aligned}
\end{equation}

\noindent Since Eq.~(\ref{eq:bound}) is bounded, we can utilize the BCT, i.e., $\lim_{\tau \to \infty} \int_{\vz^s} \nabla_{\vz^s} \mathcal{L}_{KL} \cdot d{\vz^s} = \int_{\vz^s} \lim_{\tau \to \infty} \nabla_{\vz^s} \mathcal{L}_{KL} d{\vz^s}$. Thus,

\begin{equation}
\begin{aligned}
    \lim_{\tau \to \infty} \mathcal{L}_{KL} & = \lim_{\tau \to \infty} \int_{\vz^s} \nabla_{\vz^s} \mathcal{L}_{KL} \cdot d{\vz^s} \\
    & = \int_{\vz^s} \lim_{\tau \to \infty} \nabla_{\vz^s} \mathcal{L}_{KL} d{\vz^s} \quad(\because \text{BCT})\\
    & = \int_{\vz^s} \left[ \frac{1}{K} \left( \vz^s - \vz^t \right ) - \frac{1}{K^2} \sum_{j=1}^K \left( \vz^s_j - \vz^t_j \right ) \cdot \mathbbm{1} \right] d{\vz^s} \\
    & = \frac{1}{2K} \| \vz^s - \vz^t \|_2^2  - \frac{1}{2K^2} (\sum_{j=1}^K \vz_j^s - \sum_{j=1}^K \vz_j^t)^2 + Constant \quad\quad \blacksquare
\end{aligned}
\end{equation}

In other ways, similar to the preliminary analysis, the authors showed that minimizing $\mathcal{L}_{KL}$ with sufficiently large $\tau$ is equivalent to minimizing $\mathcal{L}_{MSE}$ from $\lim_{\tau \to \infty} \frac{\partial \mathcal{L}_{KL}}{\partial \vz_k^s} = \frac{1}{K} (\vz_k^s - \vz_k^t)$ under zero-meaned logit assumption on both teacher and student. Therefore, from $\lim_{\tau \to \infty} \frac{\partial \mathcal{L}_{KL}}{\partial \vz_k^s} = \frac{1}{K} (\vz_k^s - \vz_k^t) - \frac{1}{K^2} (\sum_{j=1}^{K} \vz_k^s - \sum_{j=1}^{K} \vz_k^t)$, it is easily derived that minimizing $\mathcal{L}_{KL}$ with sufficiently large $\tau$ is equivalent to minimizing $ \left[ \mathcal{L}_{MSE} - \frac{1}{K} (\sum_{j=1}^K \vz_j^s - \sum_{j=1}^K \vz_j^t)^2 \right] $, where $\frac{1}{K}$ is a relative weight between two terms, without zero-meaned logit assumption.


\section{Detailed values of \autoref{fig:accuracy}}\label{sec:detail_value_fig1}
\autoref{tab:train_detail} and \autoref{tab:test_detail} show the detailed values in \autoref{fig:accuracy}.

\begin{table}[h]
\small\addtolength{\tabcolsep}{-2pt}
\centering
    \begin{tabular}{c|cccccccccc}
    \hline
    alpha & 0.1 & 0.2 & 0.3 & 0.4 & 0.5 & 0.6 & 0.7 & 0.8 & 0.9 & 1.0 \\
    \hline
    $\tau=1$  & 99.53 & 99.54 & 99.55 & 99.56 & 99.54 & 99.56 & 99.53 & 99.50 & 99.46 & 99.34 \\
    $\tau=3$  & 99.37 & 99.09 & 98.69 & 98.33 & 97.85 & 97.43 & 96.84 & 96.26 & 95.76 & 95.05 \\
    $\tau=5$  & 99.32 & 99.07 & 98.70 & 98.19 & 97.66 & 96.96 & 96.18 & 95.11 & 93.91 & 92.63 \\
    $\tau=20$ & 99.33 & 99.13 & 98.96 & 98.63 & 98.25 & 97.87 & 97.29 & 96.33 & 95.12 & 92.76 \\
    $\tau=\infty$ & 99.35 & 99.22 & 99.02 & 98.80 & 98.42 & 98.11 & 97.60 & 96.49 & 95.42 & 92.74 \\
    \hline
    \end{tabular}
\caption{Training accuracy on CIFAR-100 (Teacher: WRN-28-4 \& Student: WRN-16-2).}
\label{tab:train_detail}
\end{table}

\begin{table}[h]
\small\addtolength{\tabcolsep}{-2pt}
\centering
    \begin{tabular}{c|cccccccccc}
    \hline
    alpha & 0.1 & 0.2 & 0.3 & 0.4 & 0.5 & 0.6 & 0.7 & 0.8 & 0.9 & 1.0 \\
    \hline
    $\tau=1$  & 72.79 & 72.56 & 72.80 & 72.70 & 72.84 & 72.68 & 72.78 & 72.60 & 72.87 & 72.90 \\
    $\tau=3$  & 73.76 & 73.90 & 73.88 & 74.30 & 74.18 & 74.64 & 74.78 & 74.32 & 74.35 & 74.24 \\
    $\tau=5$  & 73.84 & 74.00 & 74.36 & 74.54 & 74.74 & 74.54 & 75.17 & 74.84 & 75.24 & 74.88 \\
    $\tau=20$ & 73.51 & 73.94 & 74.34 & 74.54 & 74.64 & 74.86 & 75.05 & 74.86 & 75.26 & 75.15 \\
    $\tau=\infty$ & 73.18 & 73.65 & 74.04 & 74.28 & 74.45 & 75.03 & 75.04 & 74.67 & 75.37 & 75.51 \\
    \hline
    \end{tabular}
\caption{Testing accuracy on CIFAR-100 (Teacher: WRN-28-4 \& Student: WRN-16-2).} 
\label{tab:test_detail}
\end{table}

\section{Appendix: Other methods in \autoref{tab:acc_comparison}}
We compare to the following other state-of-the-art methods from the literature:
\begin{itemize}
    \item Fitnets: Hints for thin deep nets \cite{romero2014fitnets}
    \item Attention Transfer (AT) \cite{zagoruyko2016paying}
    \item Knowledge transfer with jacobian matching (Jacobian) \cite{srinivas2018knowledge}
    \item Paraphrasing complex network: Network compression via factor transfer (FT) \cite{kim2018paraphrasing}
    \item Knowledge transfer via distillation of activation boundaries formed by hidden neurons (AB) \cite{heo2019knowledge}
    \item A comprehensive overhaul of feature distillation (Overhaul) \cite{heo2019comprehensive}
\end{itemize}

\section{Various pairs of teachers and students}\label{sec:various}

\begin{figure}[h!]
\centering
    \begin{minipage}{1.0\linewidth}
        \centering
        \begin{minipage}[t]{.49\linewidth}
            \begin{minipage}[t]{.49\linewidth}
            \centering
            \includegraphics[width=\linewidth]{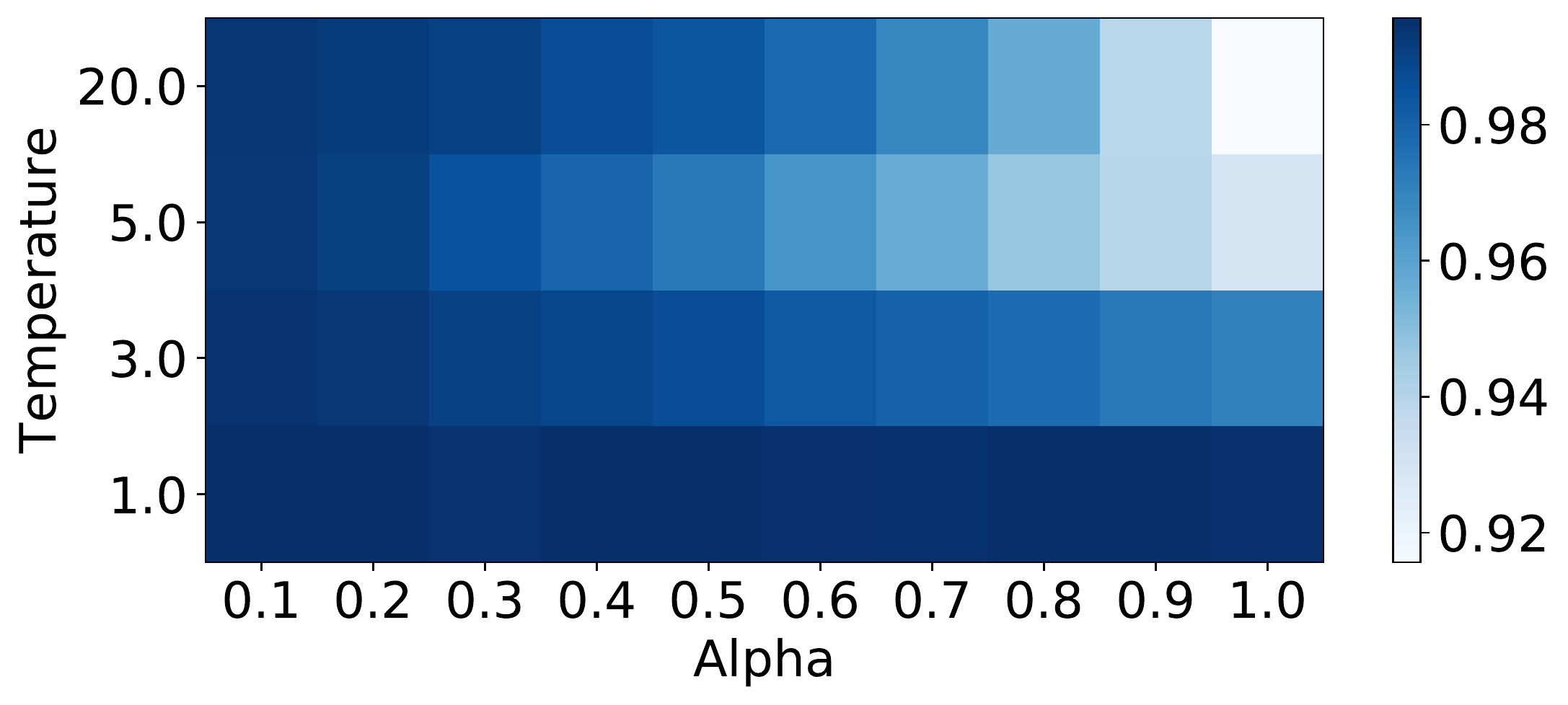}
            \end{minipage}
            \begin{minipage}[t]{.49\linewidth}
            \centering
            \includegraphics[width=\linewidth]{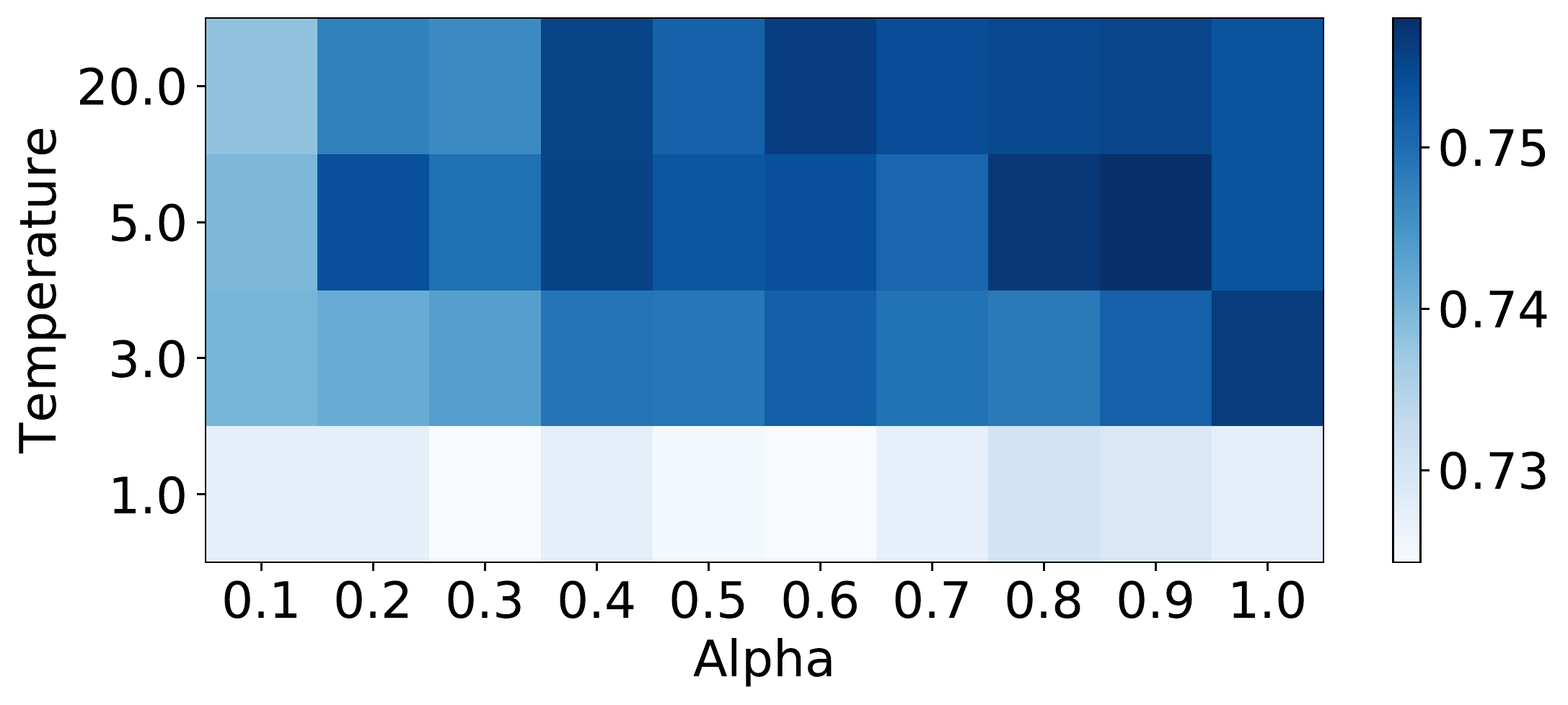}
            \end{minipage}
        \subcaption{T: WRN-16-4 \& S: WRN-16-2}
        \end{minipage}
        \begin{minipage}[t]{.49\linewidth}
            \begin{minipage}[t]{.49\linewidth}
            \centering
            \includegraphics[width=\linewidth]{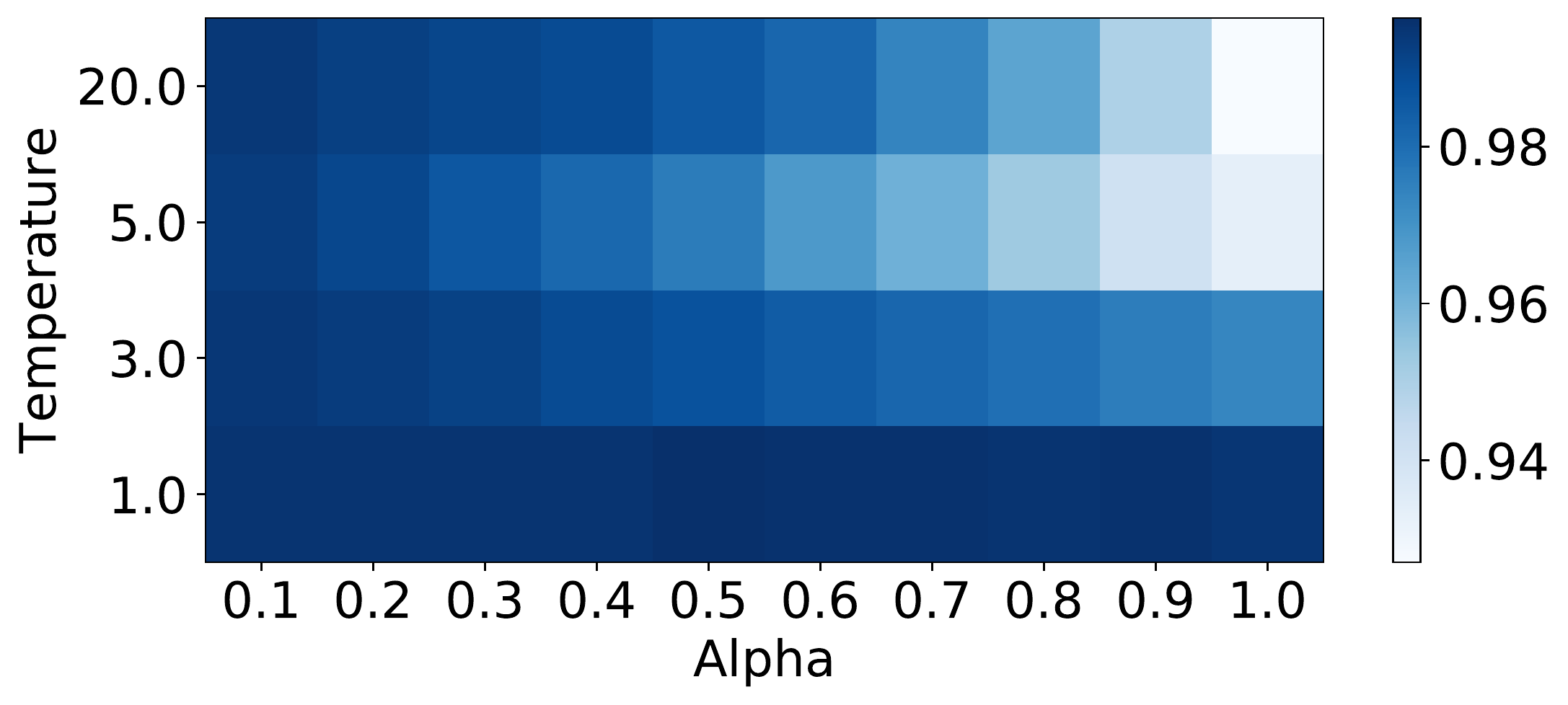}
            \end{minipage}
            \begin{minipage}[t]{.49\linewidth}
            \centering
            \includegraphics[width=\linewidth]{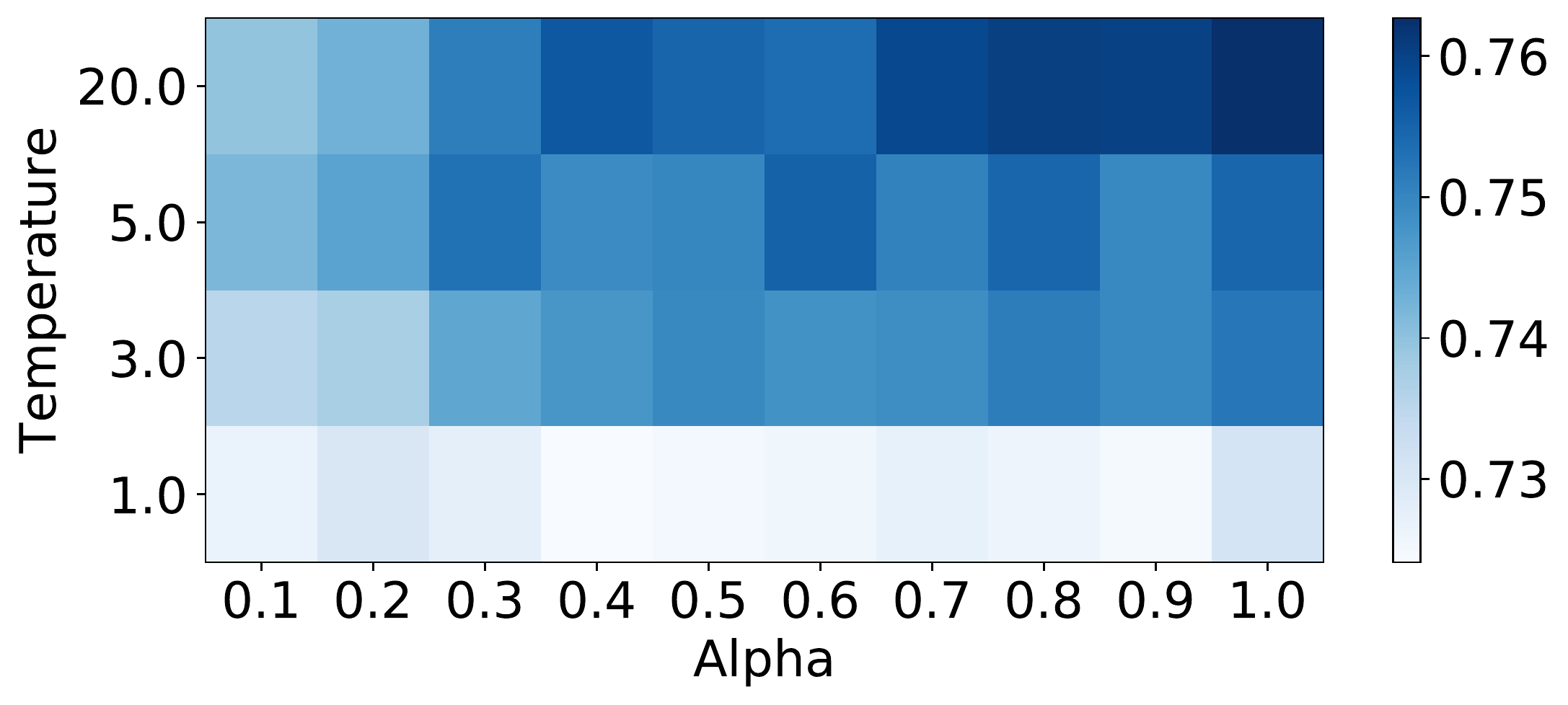}
            \end{minipage}
        \subcaption{T: WRN-16-6 \& S: WRN-16-2}
        \end{minipage}
        \\
        
        \centering
        \begin{minipage}[t]{.49\linewidth}
            \begin{minipage}[t]{.49\linewidth}
            \centering
            \includegraphics[width=\linewidth]{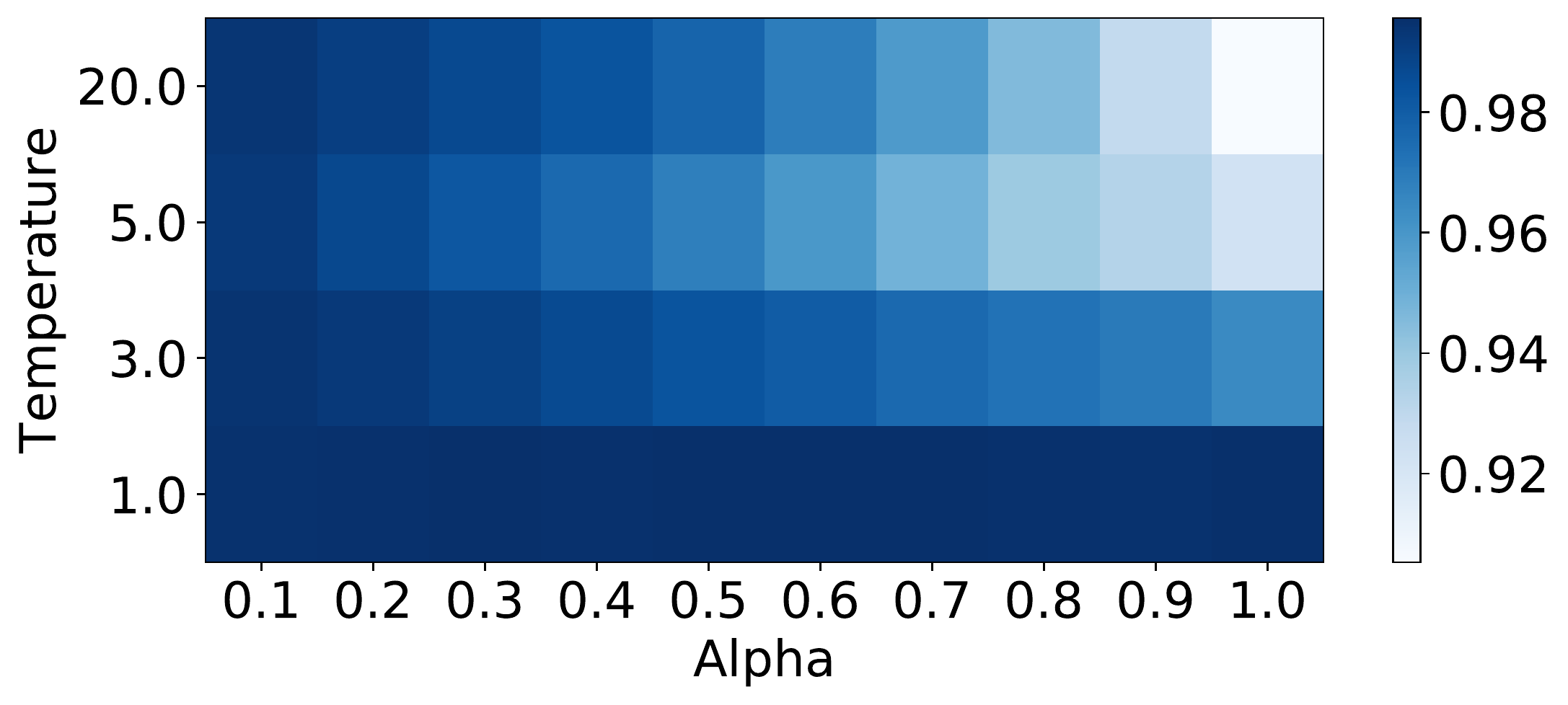}
            \end{minipage}
            \begin{minipage}[t]{.49\linewidth}
            \centering
            \includegraphics[width=\linewidth]{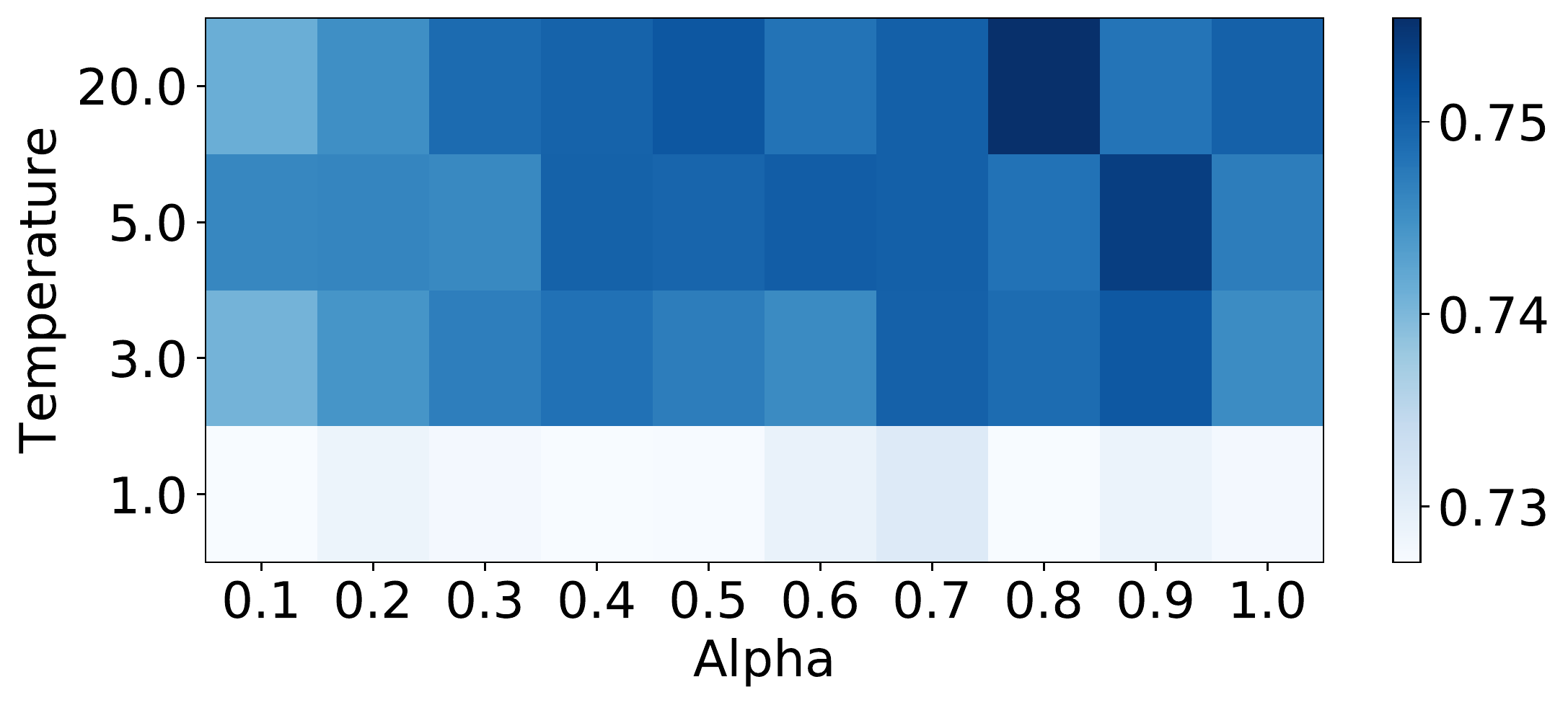}
            \end{minipage}
        \subcaption{T: WRN-28-2 \& S: WRN-16-2}
        \end{minipage}
        \begin{minipage}[t]{.49\linewidth}
            \begin{minipage}[t]{.49\linewidth}
            \centering
            \includegraphics[width=\linewidth]{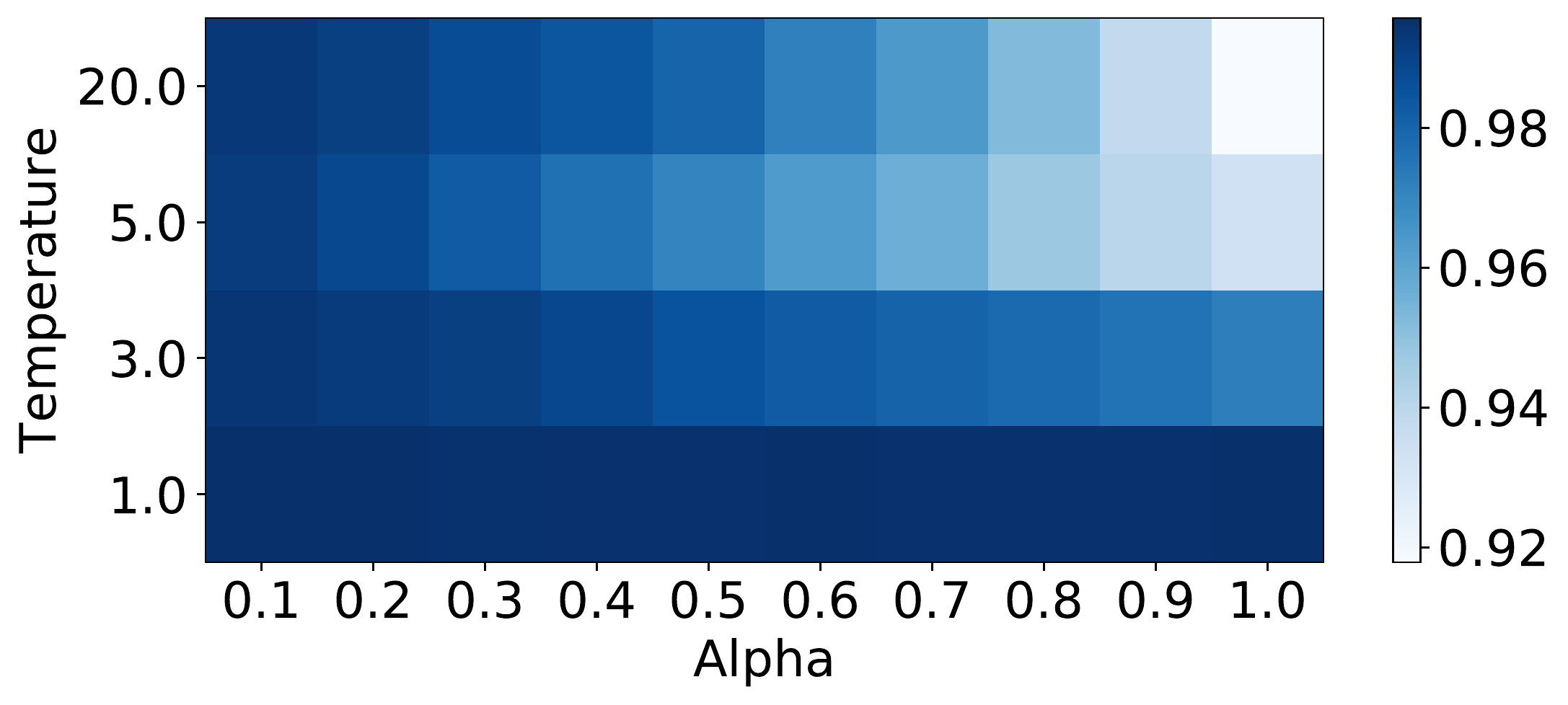}
            \end{minipage}
            \begin{minipage}[t]{.49\linewidth}
            \centering
            \includegraphics[width=\linewidth]{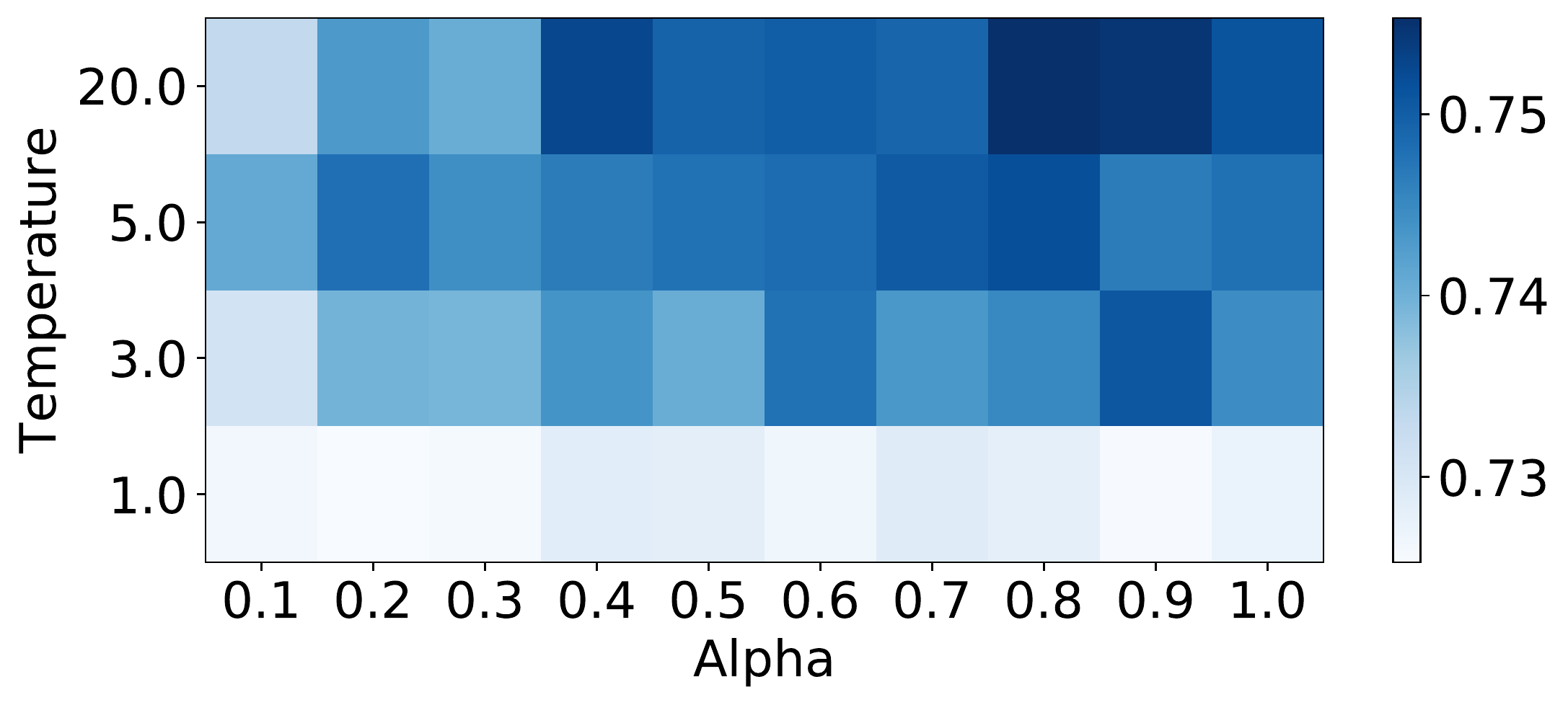}
            \end{minipage}
        \subcaption{T: WRN-40-2 \& S: WRN-16-2}
        \end{minipage}
    \end{minipage}
\caption{\textbf{Grip maps of accuracies according to the change of $\alpha$ and $\tau$} on CIFAR-100 when (a) (teacher, student) = (WRN-16-4, WRN-16-2), (b) (teacher, student) = (WRN-16-6, WRN-16-2), (c) (teacher, student) = (WRN-28-2, WRN-16-2), and (d) (teacher, student) = (WRN-40-2, WRN-16-2). The left grid maps presents training top1 accuracies, and the right grid maps presents test top1 accuracies.\label{fig:accuracy_app}}
\end{figure}

We provide the results that support the \autoref{fig:accuracy} in various pairs of teachers and students (\autoref{fig:accuracy_app}). 

\section{Calibration}
\begin{figure}[h!]
    \centering
    
    \begin{subfigure}[b]{0.24\textwidth}  
        \centering 
        \includegraphics[width=\textwidth]{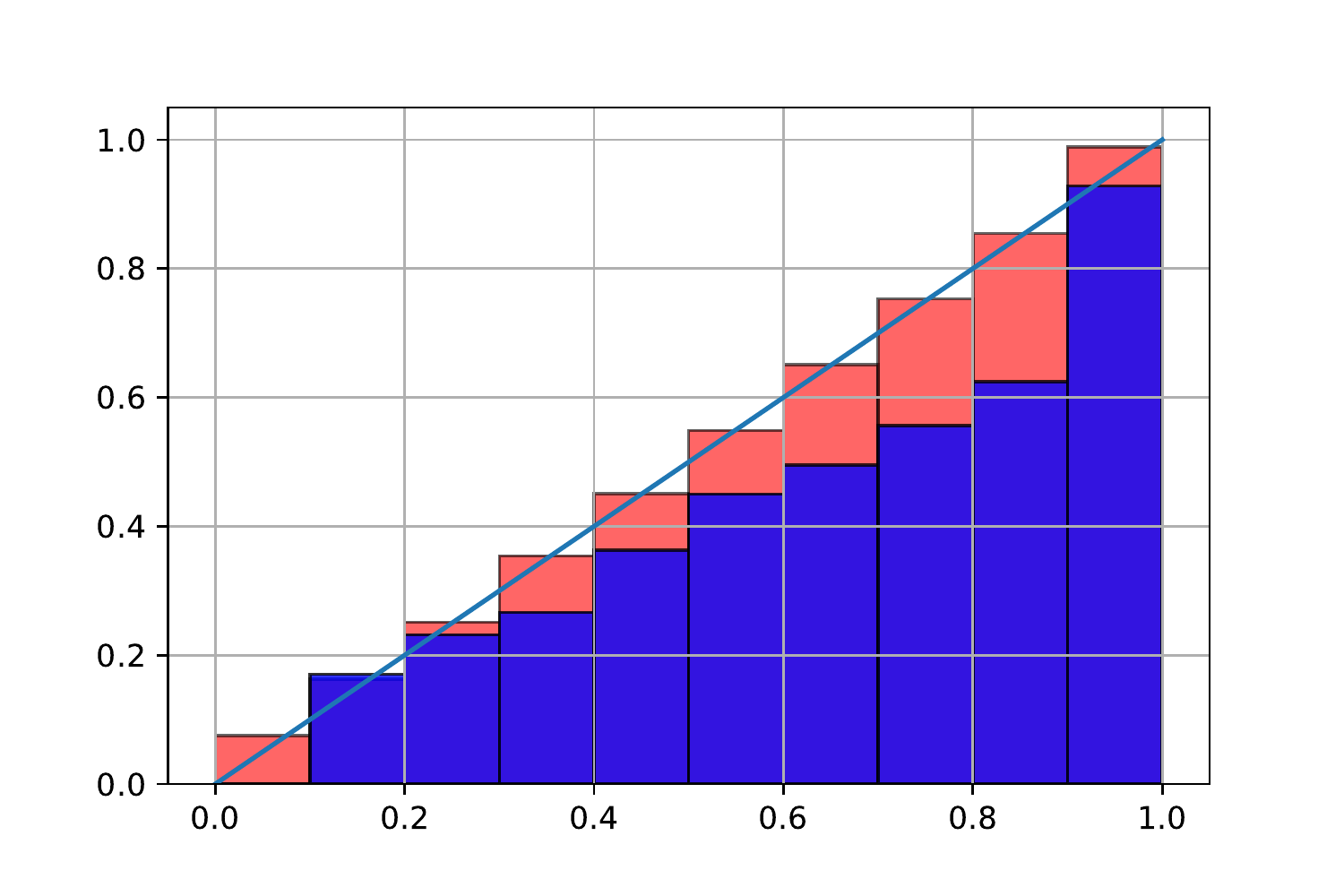}
        \caption{$t, \mathcal{L}_{CE}$ (8.44\%)}
        \label{fig5:a}
    \end{subfigure}
    \begin{subfigure}[b]{0.24\textwidth}   
        \centering 
        \includegraphics[width=\textwidth]{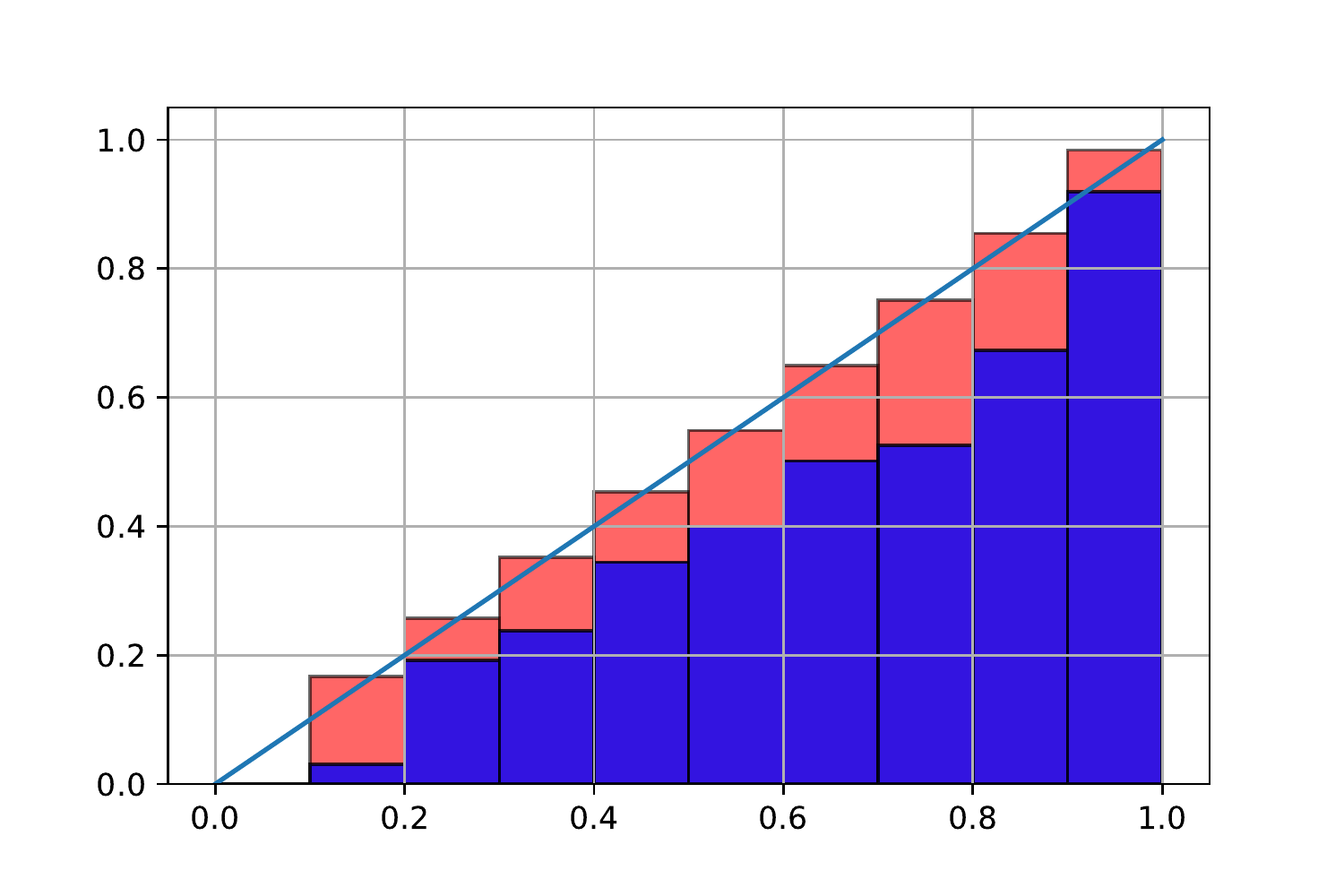}
        \caption{$s, \mathcal{L}_{CE}$ (10.33\%)}
        \label{fig5:b}
    \end{subfigure}
    \begin{subfigure}[b]{0.24\textwidth}  
        \centering 
        \includegraphics[width=\textwidth]{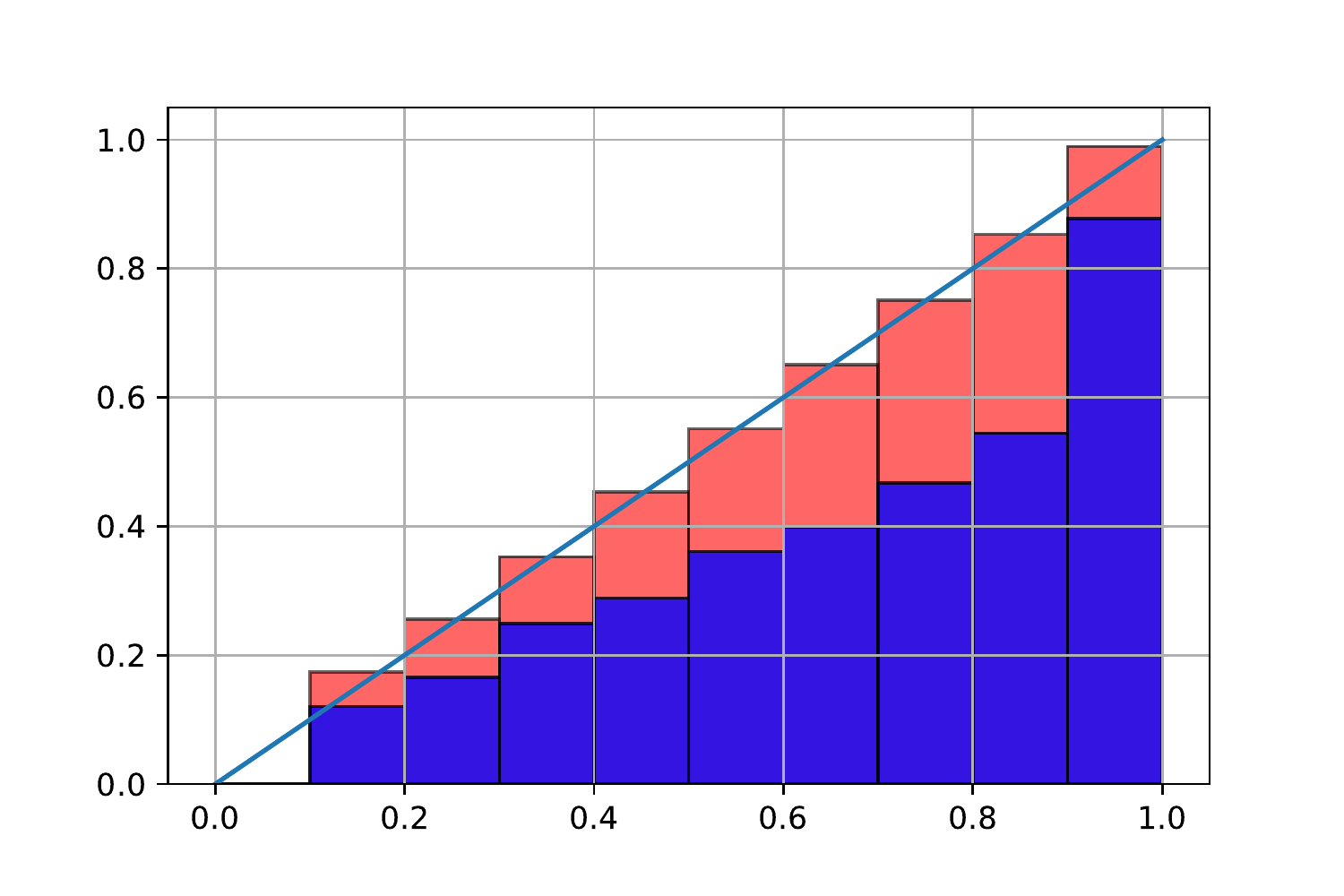}
        \caption{$s, \mathcal{L}_{KL}(\tau=\infty)$ (14.38\%)}
        \label{fig5:c}
    \end{subfigure}
    \begin{subfigure}[b]{0.24\textwidth}   
        \centering 
        \includegraphics[width=\textwidth]{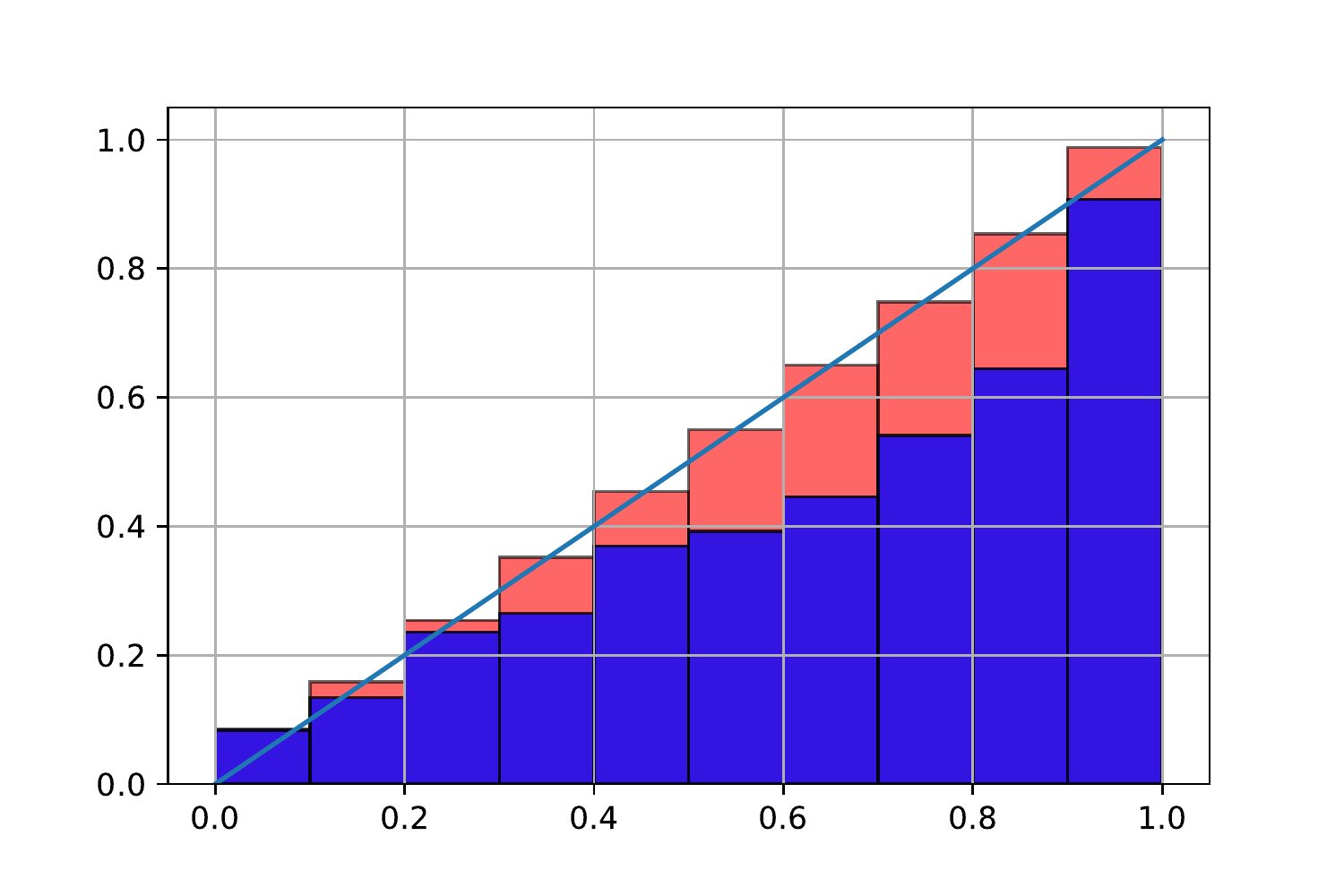}
        \caption{$s, \mathcal{L}_{MSE}$ (10.46\%)}
        \label{fig5:d}
    \end{subfigure} 
    
    
    \centering
    \caption{Reliability diagrams on CIFAR-100 training dataset. We use the (teacher, student) as (WRN-28-4 ($t$), WRN-16-2 ($s$)). Expected calibration error\,(ECE) is written in each caption\,(\%).}
    \label{fig:cali}
\end{figure}

$\delta_\infty$ encourages the penultimate layer representations to be diverged and prevents the student from following the teacher. From this observation, it is thought that $\mathcal{L}_{KL}$ and $\mathcal{L}_{MSE}$ can also have different calibration abilities. \autoref{fig:cali} shows the 10-bin reliability diagrams of various models on CIFAR-100 and the expected calibration error\,(ECE). \autoref{fig:cali}(a) and \autoref{fig:cali}(b) are the reliability diagrams of WRN-28-4 ($t$) and WRN-16-2 ($s$) trained with $\mathcal{L}_{CE}$, respectively. The slope of the reliability diagram of WRN-28-4 is more consistent than that of WRN-16-2, and the ECE of WRN-28-4 (8.44\%) is slightly less than the ECE of WRN-16-2 (10.33\%) when the models is trained with $\mathcal{L}_{CE}$. \autoref{fig:cali}(c) and \autoref{fig:cali}(d) are the reliability diagrams of WRN-16-2 ($s$) trained with $\mathcal{L}_{KL}$ with infinite $\tau$ and with $\mathcal{L}_{MSE}$, respectively. It is shown that the slope of reliability diagram of $\mathcal{L}_{MSE}$ is closer to the teacher's calibration than that of $\mathcal{L}_{KL}$. Besides, this consistency is also shown in ECE that the model trained with $\mathcal{L}_{MSE}$ (10.46\%) has less ECE than the model trained with $\mathcal{L}_{KL}$ with infinite $\tau$ (14.38\%).

\end{document}